\pdfoutput=1
\documentclass{article}

\PassOptionsToPackage{numbers, compress}{natbib}

\usepackage[main, final]{neurips_2026}

\usepackage[utf8]{inputenc}
\usepackage[T1]{fontenc}
\usepackage{url}
\usepackage{booktabs}
\usepackage{amsfonts}
\usepackage{amsmath}
\usepackage{nicefrac}
\usepackage{microtype}
\usepackage[table]{xcolor}
\usepackage{colortbl}
\definecolor{cvprblue}{rgb}{0.21,0.49,0.74}
\usepackage[pagebackref,breaklinks,colorlinks,allcolors=cvprblue]{hyperref}
\usepackage{algorithm}
\usepackage{algorithmic}
\usepackage{caption}
\usepackage{multirow}
\usepackage{adjustbox}
\usepackage{makecell}
\usepackage{graphicx}
\usepackage{rotating}
\usepackage{wrapfig}
\usepackage{pdflscape}
\usepackage{longtable}

\usepackage{placeins}
\definecolor{GroupColor1}{RGB}{220,230,255}
\definecolor{darkgreen}{HTML}{006400}
\definecolor{darkred}{HTML}{8B0000}
\newcommand{\gain}[1]{{\color{darkgreen}($\uparrow$#1)}}
\newcommand{\loss}[1]{{\color{darkred}($\downarrow$#1)}}
\newcommand{\gainci}[2]{{\color{darkgreen}($\uparrow$#1$\pm$#2)}}
\newcommand{\gainns}[2]{{\color{darkgreen}($\uparrow$#1$\pm$#2)$^{\ddagger}$}}

\usepackage{titletoc}
\usepackage{tikz}
\usetikzlibrary{decorations.pathreplacing,calc,arrows.meta,positioning}
\usepackage{xspace}

\definecolor{directcolor}{HTML}{1565C0}
\definecolor{concisecolor}{HTML}{2E7D32}
\definecolor{deepcolor}{HTML}{E65100}
\definecolor{adaptivecolor}{HTML}{6A1B9A}

\newcommand{\tagDirect}{{\color{directcolor}\texttt{<}\textsc{direct}\texttt{>}}\xspace}
\newcommand{\tagConcise}{{\color{concisecolor}\texttt{<}\textsc{concise}\texttt{>}}\xspace}
\newcommand{\tagDeep}{{\color{deepcolor}\texttt{<}\textsc{deep}\texttt{>}}\xspace}
\newcommand{\tagAdaptive}{{\color{adaptivecolor}\texttt{<}\textsc{adaptive}\texttt{>}}\xspace}

\newcommand{\modelname}{Video-FLAIR\xspace}

\usepackage{tcolorbox}

\usepackage{newfloat}
\usepackage{listings}

\newcommand{\modelwithlogo}{Video-FLAIR\xspace}

\makeatletter
\renewcommand\paragraph{\@startsection{paragraph}{4}{\z@}%
  {0.8ex \@plus0.1pt \@minus.5ex}%
  {-0.2em}%
  {\normalfont\normalsize\bfseries}}
\makeatother
\makeatletter

\title{\modelname: Not Whether to Reason, But How}

\author{
    Yogesh Kulkarni \quad Pooyan Fazli \\
    Arizona State University \\
    {\tt\small \{ykulka10, pooyan\}@asu.edu}\\
    {\tt \small \href{https://people-robots.github.io/Video-FLAIR/}{\textcolor{black}{\texttt{https://people-robots.github.io/Video-FLAIR/}}}}
}

\begin{document}

\maketitle
\begin{abstract}

Multimodal queries can require different types of reasoning. Some can be answered via \emph{perceptual} reasoning, extracting information directly from the visual signal, while others require \emph{compositional} reasoning that combines observations or \emph{deliberative} reasoning that evaluates competing hypotheses. However, many existing methods apply a uniform reasoning strategy across queries, leading to unnecessary computation on simple tasks and insufficient reasoning on complex ones.\ We introduce \textbf{\modelname}, a training framework that learns to select the appropriate reasoning mode for each query using reinforcement learning. During training, the model generates responses under all three modes for the same prompt, enabling direct comparison. A composite reward compares these responses to favor the most effective one based on correctness, grounding, and cost, while discouraging unsupported or misaligned deliberation. This yields a supervision signal for learning adaptive reasoning without per-query annotations. \modelname improves accuracy over the Qwen2.5-VL base model by $\mathbf{+5.4}$ on MathVista, $\mathbf{+4.8}$ on Video-Holmes, and $\mathbf{+4.8}$ on Video-MMMU, while reducing average token usage to $\mathbf{95}$ compared to $417$ for always-thinking baselines.

\end{abstract}

\section{Introduction}

Multimodal LLMs achieve strong benchmark performance~\citep{sui2025stop, videopasta}, but they struggle to adapt their reasoning to each query. Applying chain-of-thought (CoT) reasoning uniformly across queries uses up to $20{\times}$ more tokens than needed for simple queries~\citep{chen2025donot,su2025between}. As reasoning chains grow longer, models can shift attention away from visual evidence and produce hallucinated intermediate steps~\citep{tian2025more}. At the same time, complex queries may fail to receive the multi-step or hypothesis-driven reasoning required to answer them. The root problem is that current methods either apply the same reasoning strategy across queries~\cite{feng2025videor1, maaz2025videor2, wang2025videorft,videosavi} or decide only whether to use reasoning at all~\cite{liu2026videoautor1}, without considering what kind of reasoning a query requires.
For instance, some queries (e.g., reading on-screen text or identifying an object) can be answered by directly extracting information from the visual signal. Others (e.g., localizing an event across video frames or computing a quantity from chart data) require combining observations through a short, grounded sequence of steps~\citep{wei2022chain}. More demanding queries (e.g., evaluating whether an athlete’s behavior indicates confidence) require reasoning that evaluates and rejects competing interpretations.

\begin{figure*}
    \centering
    \includegraphics[width=\textwidth]{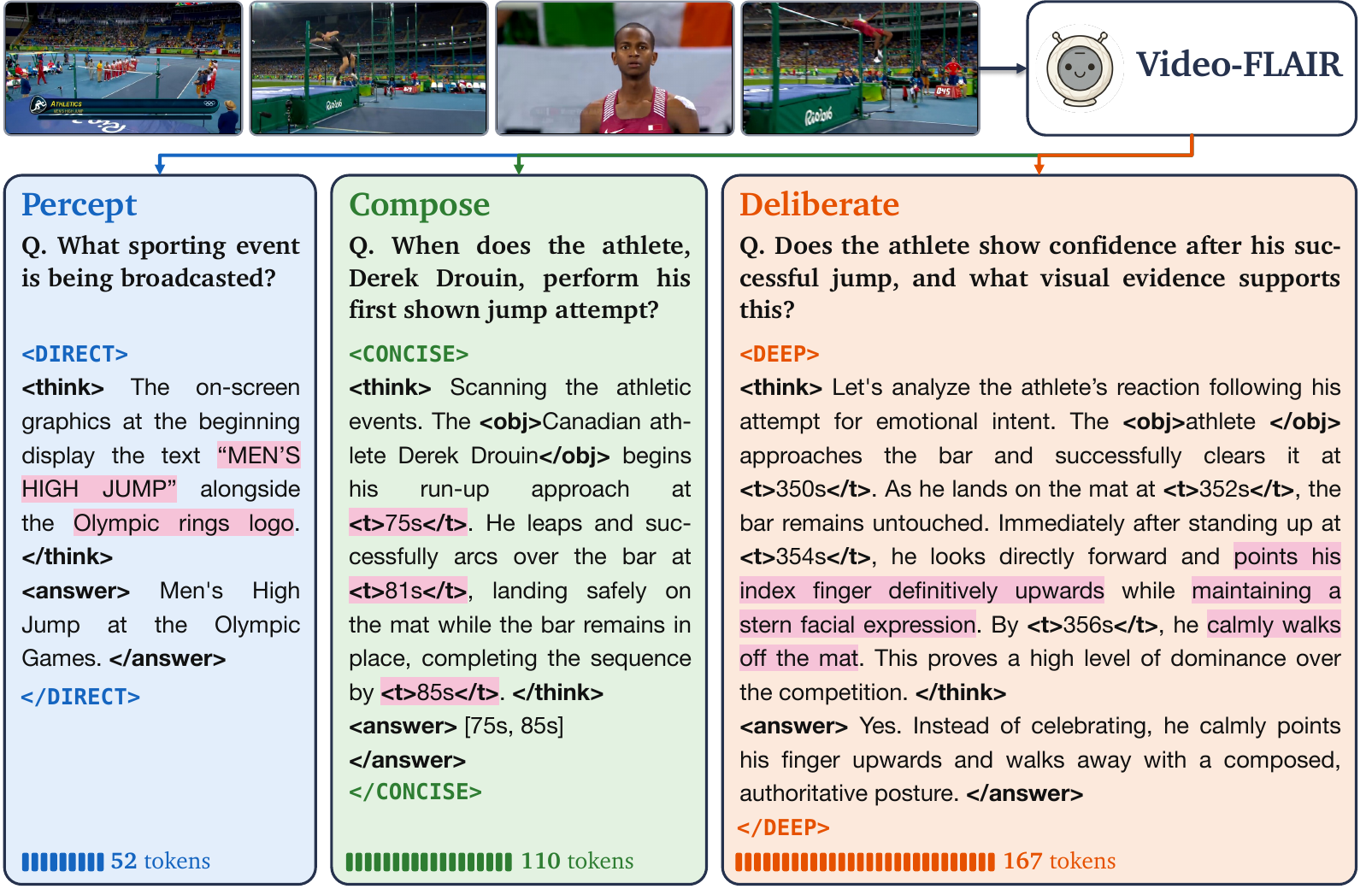}
\caption{\textbf{\modelname} routes each query to one of three reasoning modes based on its complexity. Key evidence driving each mode is shown in {\setlength{\fboxsep}{1pt}\colorbox[HTML]{F6C3D8}{pink}}, namely screen-text reads ({\color{directcolor}\texttt{<}\textsc{direct}\texttt{>}}), timestamped event anchors ({\color{concisecolor}\texttt{<}\textsc{concise}\texttt{>}}), and multi-observation cues synthesized into a conclusion ({\color{deepcolor}\texttt{<}\textsc{deep}\texttt{>}}). The counter on each panel gives the number of reasoning tokens, which grows only when the query requires deeper reasoning.}
    \label{fig:teaser}
\end{figure*}

These differences reflect distinct types of reasoning and suggest that reasoning should be treated as an \textit{adaptive} process rather than a fixed strategy. We therefore define three reasoning modes: \textsc{Percept} (\tagDirect), \textsc{Compose} (\tagConcise), and \textsc{Deliberate} (\tagDeep). \textsc{Percept} maps \textit{directly} from evidence to answer, \textsc{Compose} builds answers through a \textit{concise} sequence of observations, and \textsc{Deliberate} revisits the same evidence to reason \textit{deeply} about competing hypotheses.

Building on this formulation, we introduce \textbf{\modelname} (\textbf{F}lexible \textbf{L}earned \textbf{A}daptive \textbf{I}nference and \textbf{R}easoning), a training framework that uses reinforcement learning to select an appropriate reasoning mode for each query. During training, the model produces responses under all three modes for the same prompt, allowing their outcomes to be directly compared. A composite reward then favors the most effective mode based on correctness, grounding, and cost, while discouraging unsupported or misaligned deliberation. This provides a supervision signal without per-query annotations.

In summary, our contributions are as follows:

\begin{enumerate}

\item We introduce \modelname, a training framework for adaptive multimodal reasoning that selects among \textsc{Percept}, \textsc{Compose}, and \textsc{Deliberate} modes at inference, enabling models to learn \emph{how to reason} rather than applying a fixed strategy.

  \item We propose mode-structured rollouts that generate responses under all three reasoning modes for the same query, enabling within-query comparison. This framework is guided by a composite reward that includes an online verifier to promote grounded reasoning.

    \item \modelname improves accuracy by $+5.4$ on MathVista, $+4.8$ on Video-Holmes, and $+4.8$ on Video-MMMU over Qwen2.5-VL, while reducing average token usage to $95$ compared to $417$ (across models and benchmarks) for baselines that apply an always-thinking strategy.

\end{enumerate}

\section{Related Work}

\paragraph{Overthinking and over-reasoning in Multimodal LLMs.}

Chain-of-thought reasoning is not uniformly beneficial~\citep{chen2025towards,sui2025stop}. Models can use up to 20$\times$ more tokens than necessary on simple queries~\citep{chen2025donot,su2025between}, and excessive reasoning can degrade performance when the task does not require deliberation~\citep{tian2025more,fan2025missing,kang2025c3ot}. In vision-language models, long reasoning chains can also shift attention away from visual input, leading to progressive visual forgetting and hallucinated intermediate steps~\citep{tian2025more, egovita}, while substantially increasing token costs~\citep{kumar2025overthink}.
We address this by learning a cost-aware policy that selects among different reasoning modes, rather than applying a uniform or binary reasoning strategy.

\paragraph{Adaptive and budget-aware reasoning.}
Prior work on adaptive reasoning focuses on binary mode switching, deciding whether to invoke reasoning or not. For example, AdaptThink~\citep{zhang2025adaptthink} trains a think/no-think selector that reduces response length by 53\%, KAT-V1~\citep{zhan2025katv1} uses trigger-token-based switching, R-4B~\citep{yang2025r4b} introduces bi-mode annealing, and Video-Auto-R1~\citep{liu2026videoautor1} applies confidence-based early exit for video QA. Similar binary strategies extend to audio~\citep{wu2025audiothinker}, omni-modal~\citep{yang2025omniautothink, avatar}, and Pareto-optimized settings~\citep{lou2025adacot}, while AdaTooler-V~\citep{wang2025adatoolerv} highlights how naive tool use can worsen overthinking.
Beyond binary switching, ARM2~\citep{xie2025arm2} expands the action space to multiple reasoning formats, but relies on fixed SFT-defined templates rather than learning how to select among them.

\paragraph{Reinforcement learning-based reasoning.}
A parallel line of work improves reasoning in multimodal models through reinforcement learning, including Video-R1~\citep{feng2025videor1}, Video-R2~\citep{maaz2025videor2}, and VideoRFT~\citep{wang2025videorft}. While effective, these approaches continue to apply a single reasoning style across queries, without adapting the reasoning structure to the task. \modelname differs by learning how to select among reasoning modes through structured rollouts that compare alternative strategies on the same prompt.

\section{Preliminaries}
\subsection{Adaptive Reasoning Space}
\label{sec:prelim}
\label{sec:reasoning_space}

Various cognitive frameworks suggest that different tasks impose qualitatively different demands, motivating distinct reasoning strategies~\citep{bloom1956taxonomy,hattie2016learning}. We define three reasoning modes:

\begin{itemize}
\item \textsc{Percept} (\tagDirect): Perceptual reasoning extracts a single observation that maps \emph{directly} to the answer. The evidence chain has no branches, and the reasoning path is immediate.

\item \textsc{Compose} (\tagConcise): Compositional reasoning combines multiple observations through a linear sequence, where each step builds on the previous one without revisiting or branching. The structure is \emph{concise} in that the chain is bounded rather than short.

\item \textsc{Deliberate} (\tagDeep): Deliberative reasoning revisits the same evidence to evaluate and reject competing hypotheses. The model reasons \emph{deeply} by exploring alternatives within the same evidence, so depth reflects search over interpretations rather than longer chains.
\end{itemize}

\section{\texorpdfstring{\modelname}{Video-FLAIR}}
\label{sec:method}
\begin{figure*}
    \centering
    \includegraphics[width=\textwidth]{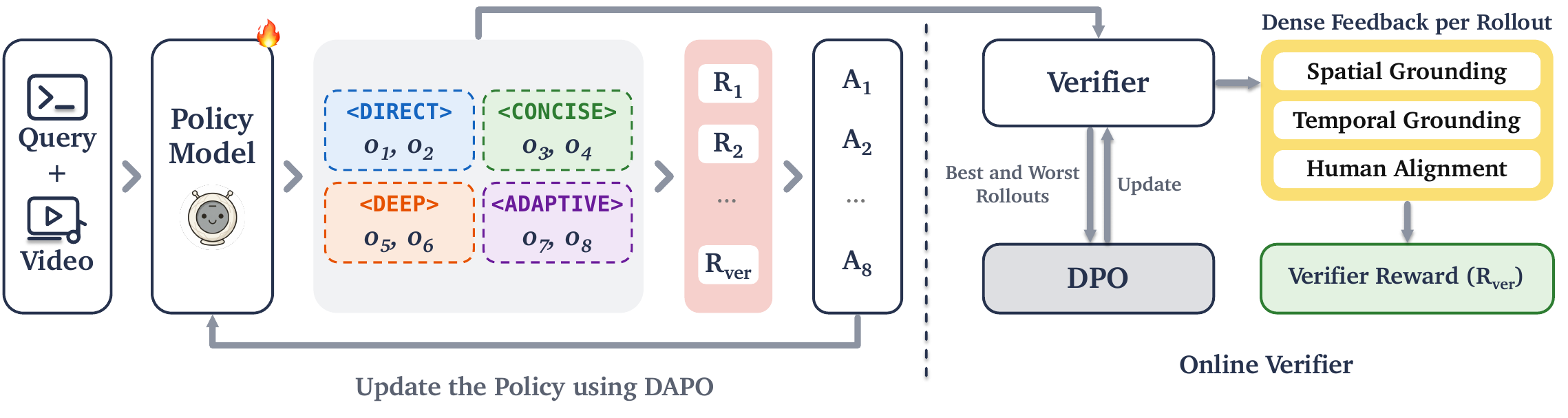}
    \caption{\textbf{\modelname training pipeline.} \textbf{(Left)} Each video-query pair generates $K{=}8$ mode-structured rollouts: two controlled rollouts per mode ({\color{directcolor}\texttt{<}\textsc{direct}\texttt{>}}, {\color{concisecolor}\texttt{<}\textsc{concise}\texttt{>}}, {\color{deepcolor}\texttt{<}\textsc{deep}\texttt{>}}) and two {\color{adaptivecolor}\texttt{<}\textsc{adaptive}\texttt{>}}. A composite reward is computed for each rollout. GDPO~\cite{liu2026gdpo} normalizes each reward component before aggregation. The policy is then updated via DAPO~\cite{yu2025dapo}. \textbf{(Right)} The verifier provides per-rollout feedback aggregated into $R_{\mathrm{ver}}$. It is periodically refreshed via DPO~\cite{dpo} on highest- versus lowest-reward rollout pairs to remain calibrated with the improving policy.}
    \label{fig:pipeline}
\end{figure*}

\modelname trains a multimodal LLM policy to select a reasoning mode per query using reinforcement learning. Starting from an SFT warm-start (\S\ref{sec:sft}), the model generates mode-structured rollouts that evaluate all three reasoning modes on the same prompt (\S\ref{sec:forced_slots}). These rollouts are scored using a composite reward (\S\ref{sec:reward}) that incorporates an online verifier for dense grounding feedback (\S\ref{sec:verifier_guided}). A utility function derived from this reward identifies the most effective mode and supervises the adaptive slots (\S\ref{sec:utility}).
The resulting advantages are normalized via GDPO and applied token-wise (\S\ref{sec:advantage}) before updating the policy with DAPO (\S\ref{sec:dapo}).

\subsection{SFT Warm-Start}
\label{sec:sft}

We warm-start the model with supervised fine-tuning on mode-structured traces, ensuring that the subsequent RL stage focuses on reasoning rather than format learning. All outputs follow a strict schema: \texttt{<MODE><think>...</think><answer>...</answer></MODE>}, where \texttt{MODE} is one of \tagDirect, \tagConcise, or \tagDeep. Reasoning traces for \tagConcise and \tagDeep may include structured evidence tokens: \texttt{<obj>...</obj>} for referenced entities, \texttt{<box>...</box>} for spatial regions, and \texttt{<t>...</t>} for temporal timestamps. Dataset details are provided in Appendix~\S\ref{app:data_construction}.

\subsection{Mode-Structured Rollout Groups}
\label{sec:forced_slots}

Selecting the appropriate reasoning mode requires comparing outcomes across modes on the same input under a shared reward. We therefore use \emph{mode-structured rollouts}, generating two responses under each reasoning mode for the same (visual input, query) pair using mode-specific prompt templates (Figures~\ref{fig:prompt_direct}, \ref{fig:prompt_concise}, \ref{fig:prompt_deep}, \ref{fig:prompt_adaptive}). SFT ensures the model follows each mode correctly, so reward differences within each rollout group reflect which mode is most effective for the input. For each (visual input, query) pair, we sample $K=8$ rollouts with a deterministic slot assignment:

\vspace{-0.4cm}
\begin{equation}
\label{eq:slot_order}
  [\underbrace{\tagDirect,\tagDirect}_{\text{slots 1--2}},\,
   \underbrace{\tagConcise,\tagConcise}_{\text{slots 3--4}},\,
   \underbrace{\tagDeep,\tagDeep}_{\text{slots 5--6}},\,
   \underbrace{\tagAdaptive, \tagAdaptive}_{\text{slots 7--8}}].
\end{equation}
\vspace{-0.4cm}

Slots 1--6 are \emph{controlled}: the prompt explicitly specifies the reasoning mode, forcing the model to generate responses under \tagDirect, \tagConcise, or \tagDeep.
Slots 7--8 are \emph{adaptive}: the prompt provides definitions of all reasoning modes without specifying one, requiring the model to select a mode before generating its response.

\subsection{Composite Reward}
\label{sec:reward}

Each rollout is scored by a composite reward:
\begin{align}
  \label{eq:composite_reward}
  R = \, & w_{\mathrm{ans}} R_{\mathrm{ans}}
           + w_{\mathrm{format}} R_{\mathrm{format}}
           + w_{\mathrm{comply}} R_{\mathrm{comply}} \notag \\
          &+ w_{\mathrm{cost}} R_{\mathrm{cost}}
           + w_{\mathrm{balance}} R_{\mathrm{balance}}
           + w_{\mathrm{select}} R_{\mathrm{select}}
           + w_{\mathrm{verifier}} R_{\mathrm{verifier}},
\end{align}
Appendix Table~\ref{tab:hyperparams} lists the weights, and Appendix~\S\ref{app:reward_math} provides full definitions. The seven terms form three functional groups: (1) correctness and validity, (2) cost efficiency, and (3) adaptive selection.

\paragraph{Correctness and validity ($R_{\mathrm{ans}}$, $R_{\mathrm{format}}$, $R_{\mathrm{comply}}$).}
$R_{\mathrm{ans}}$ evaluates task correctness based on the answer type (Appendix~\S\ref{app:reward_ans}).
$R_{\mathrm{format}}$ equals $1$ when the response follows the required tag schema (\texttt{<MODE><think>...</think><answer>...</answer></MODE>}), and $0$ otherwise.
$R_{\mathrm{comply}}$ penalizes rollouts that either violate the assigned mode instruction (e.g., a forced \tagDirect slot producing a \tagDeep response) or exceed the word budget for that mode.

\begin{wrapfigure}{r}{0.39\linewidth}
  \centering
  \vspace{-16pt}
  \includegraphics[width=\linewidth]{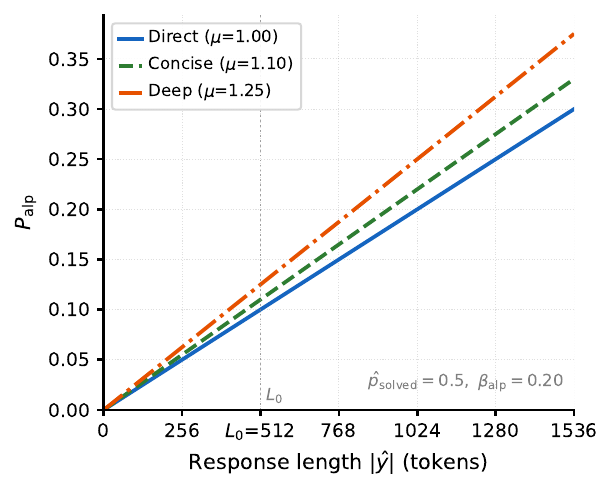}
  \caption{\textbf{Penalty scales with length.}}
  \label{fig:alp_scaling}
  \vspace{-3pt}
\end{wrapfigure}

\paragraph{Cost efficiency ($R_{\mathrm{cost}}$, $R_{\mathrm{balance}}$).} $R_{\mathrm{cost}}$ penalizes excessive reasoning while adapting to query difficulty. It combines a base cost term and an adaptive length penalty:
\begin{equation}
R_{\mathrm{cost}} = -(P_{\mathrm{base}} + P_{\mathrm{alp}}),
\end{equation}
where $P_{\mathrm{base}}$ accounts for fixed per-mode costs and any word-budget overflow (Appendix~\S\ref{app:reward_math}), and $P_{\mathrm{alp}}$ is the Adaptive Length Penalty (ALP)~\citep{alp}:
\begin{equation}
\label{eq:alp}
P_{\mathrm{alp}} = \beta_{\mathrm{alp}} \cdot \frac{|\hat{y}|}{L_0} \cdot \hat{p}_{\mathrm{solved}} \cdot \mu_m,
\end{equation}

where $\hat{p}_{\mathrm{solved}}$ is the fraction of correct rollouts across all $K{=}8$ slots, and $\mu_m$ is a per-mode multiplier (Appendix~\S\ref{app:reward_math}, Figure~\ref{fig:alp_scaling}). We extend ALP with the per-mode multiplier $\mu_m$, which differentiates cost pressure across modes, and with difficulty-aware scaling through $\hat{p}_{\mathrm{solved}}$. Because the rollout group includes \tagDirect and \tagConcise controlled slots, $\hat{p}_{\mathrm{solved}}$ serves as a proxy for query difficulty: it is high on easy queries and low on hard ones. As a result, cost penalties are stronger for simple queries, discouraging unnecessary deliberation, and weaker for complex queries, allowing deeper reasoning when needed. This ensures that \tagDeep is not penalized when it is necessary for solving the task (see Appendix~\S\ref{app:cost_analysis} for a case analysis). $R_{\mathrm{balance}}$ encourages diversity across reasoning modes by penalizing imbalanced mode usage within each rollout group.

\paragraph{Adaptive selection ($R_{\mathrm{select}}$, $R_{\mathrm{verifier}}$).}
$R_{\mathrm{select}}$ applies utility-derived supervision to the \tagAdaptive slots (\S\ref{sec:utility}), while $R_{\mathrm{verifier}}$ provides dense grounding and alignment feedback (\S\ref{sec:verifier_guided}). Together, they determine \emph{which reasoning mode to select} and \emph{how well its outputs are grounded in the input}.

\subsection{Learning to Be Adaptive: Utility and Selection}
\begin{wrapfigure}{r}{0.35\linewidth}
  \centering
  \vspace{-16pt}
  \includegraphics[width=\linewidth]{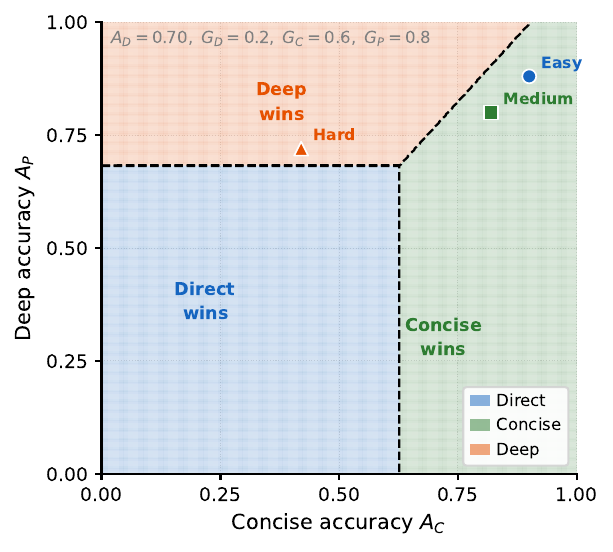}
 \caption{\textbf{Which mode wins?} Decision regions over $(A_C, A_P)$. More expensive modes win only when their accuracy-grounding gain exceeds the cost
  threshold.}
  \label{fig:utility_regions}
  \vspace{-5pt}
\end{wrapfigure}

\label{sec:utility}

\smallskip
We compute a utility score within each rollout group to determine the most effective reasoning mode.

\textbf{Step 1: Score controlled rollouts.}
For each reasoning mode $m$, we compute two quantities from its controlled rollouts: $\mathrm{Ans}(m)$, the mean correctness derived from $R_{\mathrm{ans}}$, and $G(m) = \tfrac{1}{2}(v_t(m) + v_s(m))$, where $v_t(m), v_s(m) \in [0,1]$ are the verifier's temporal and spatial grounding scores for mode $m$ (Appendix~\S\ref{app:verifier_interface}).

\textbf{Step 2: Compute utility and select the optimal mode.} We define the utility of each mode as:
\begin{equation}
  \label{eq:utility}
  \mathrm{Utility}(m) = \bigl(\mathrm{Ans}(m) + \delta \cdot G(m)\bigr) \cdot \bigl(1 - \mathrm{Cost}(m)\bigr).
\end{equation}

The weight $\delta$ is tuned so that grounding resolves near-ties in accuracy, while preventing a grounded \emph{incorrect} answer from exceeding an accurate ungrounded one (sensitivity analysis in Table~\ref{tab:sensitivity}). The $\mathrm{Cost}(m)$ term includes the per-mode cost $C_m$ and word-count overflow, penalizing deeper modes even within budget (cost-pressure analysis in \S\ref{app:cost_analysis}). We select the optimal mode as $m^{*} = \arg\max_m \mathrm{Utility}(m)$. When accuracy and grounding are equal, the cost term alone determines the ranking, favoring the more efficient mode (Figure~\ref{fig:utility_regions}).

\textbf{Step 3: Assign $R_{\mathrm{select}}$ to adaptive slots.}
Adaptive slots are evaluated against $m^{*}$. If an \tagAdaptive slot selects $m^{*}$, it receives a reward bonus. Otherwise, it receives a symmetric penalty.

\subsection{Verifier-Guided Dense Feedback}
\label{sec:verifier_guided}

We use Qwen3-VL-8B~\citep{qwen3vl} as an online verifier that evaluates each reasoning trace and produces three signals: temporal grounding $v_t$ (alignment between timestamps/event ordering and visual content), spatial grounding $v_s$ (alignment between referenced regions or objects and visual content), and human alignment $v_h$ (consistency between the reasoning trace and observable evidence). These are the main continuous inputs to $R_{\mathrm{verifier}}$, a component of the composite reward; the full expression including span and mode-alignment terms is in Appendix~\S\ref{app:reward_ver} (system prompt in Fig.~\ref{fig:prompt_verifier}).
The verifier is initialized from the base Qwen3-VL-8B checkpoint rather than the SFT policy, runs as a separate instance, and receives no gradients from the policy loss. \tagDirect completions that introduce grounding tags are penalized to prevent mode collapse, ensuring that structured grounding is used only when reasoning requires it. Because \tagDirect rollouts carry no grounding tags, they receive low $v_t$ and $v_s$ by design. The verifier instead evaluates them through $v_h$, the tag penalty, and its preferred-mode prediction, and GDPO normalizes each reward component independently (\S\ref{sec:advantage}), so \tagDirect still wins on easy perceptual queries through its lower cost (Table~\ref{tab:utility_cases}). \tagAdaptive slots are also penalized when their reasoning contradicts the verifier’s grounding assessment.

To prevent verifier drift, we periodically refresh the verifier using preference optimization. Every $N=100$ RL steps, we construct preference pairs $\mathcal{P}_{\mathrm{pref}}$ from the highest- and lowest-reward rollouts within each group and fine-tune the verifier $V$ via Direct Preference Optimization (DPO)~\cite{dpo}, producing an updated verifier $V'$ that is hot-swapped into the reward pipeline (Appendix~\S\ref{app:verifier_dpo}). Rollouts are ranked by the composite reward, in which $R_{\mathrm{ans}}$ dominates, so a rollout with high verifier scores but a wrong answer is still rejected. Each refresh uses at most $2{,}000$ pairs with a reward margin above $0.15$, and the refreshed verifier retains its general capability (Table~\ref{tab:verifier_drift}). The refresh loop is
\[
  \pi_\theta \,\xrightarrow{\text{rollouts}}\, \mathcal{P}_{\mathrm{pref}} \,\xrightarrow{\text{DPO}}\, V' \,\xrightarrow{\text{hot-swap}}\, R_{\mathrm{verifier}},
\]
where $\pi_\theta$ is the current policy being trained.

\subsection{Advantage Shaping}
\label{sec:policy_update}

\paragraph{GDPO.}
\label{sec:advantage}

Standard GRPO normalization can collapse multiple reward components into identical advantage values, obscuring distinctions between signals such as $R_{\mathrm{select}}$ and $R_{\mathrm{cost}}$. GDPO~\citep{liu2026gdpo} addresses this by normalizing each component independently before aggregation.
Each reward component $r_k$ is normalized within its rollout group:
\begin{equation}
  \label{eq:gdpo}
  A_k^{(i,j)} = \frac{r_k^{(i,j)} - \mathrm{mean}_j\{r_k^{(i,j)}\}}{\mathrm{std}_j\{r_k^{(i,j)}\}+\epsilon},
  \qquad
  A_{\mathrm{sum}}^{(i,j)} = \sum_{k=1}^{7} A_k^{(i,j)},
\end{equation}
where $i$ indexes input-query pairs and $j$ indexes rollouts within each group. The aggregated advantage $A_{\mathrm{sum}}$ is then batch-renormalized so each reward component contributes a properly scaled gradient.

\paragraph{Token-Level Credit Shaping.}
\label{sec:token_credit}

\begin{wrapfigure}{r}{0.38\linewidth}
\centering
\vspace{-13pt}
\includegraphics[width=\linewidth]{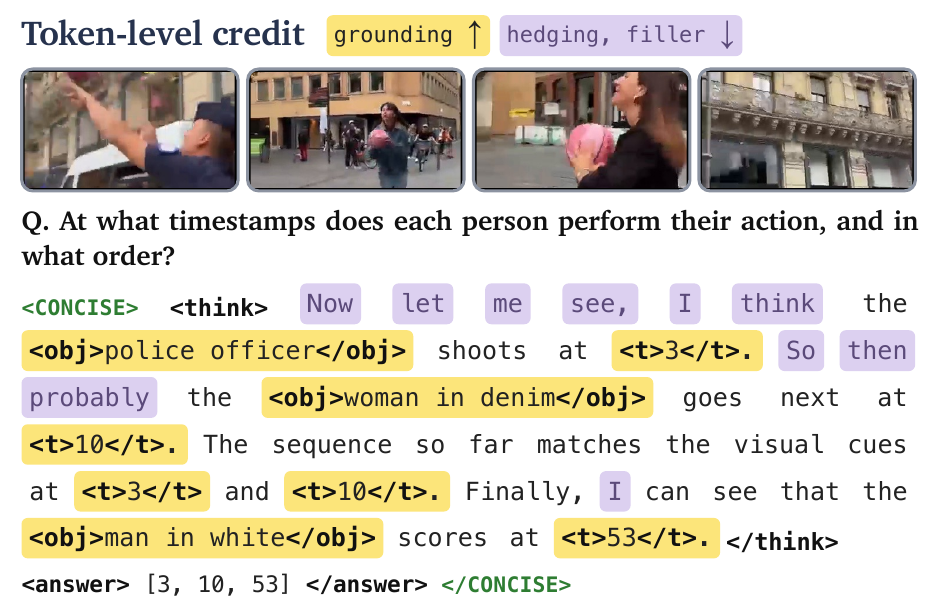}
\caption{%
  \textbf{Token-level credit assignment.}
  Yellow spans (grounding tags, factual observations) receive amplified gradients, while purple spans (hedging, filler) are suppressed.
}
\label{fig:token_credit}
\vspace{-16pt}
\end{wrapfigure}

The batch-renormalized $A_{\mathrm{sum}}$ from GDPO is a single scalar per rollout.\ Token-level credit shaping redistributes this signal across tokens, concentrating gradient updates on evidence-bearing spans rather than applying uniform weighting across the completion. We assign each token a weight $c_t$ and apply it to $A_{\mathrm{sum}}$:
\begin{equation}
  \label{eq:token_credit}
  \hat{A}_t = A_{\mathrm{sum}} \cdot c_t, \qquad \frac{1}{T}\sum_{t=1}^{T} c_t = 1,
\end{equation}
where $T$ is the completion length and $c_t \geq 0$.

Spans are classified using tag matching and verifier trace annotations (Appendix~\S\ref{app:token_credit}).
\emph{Evidence-bearing} spans receive higher weights, including tokens inside \texttt{<t>}, \texttt{<obj>}, and \texttt{<box>} tags, as well as spans identified by the verifier as grounded in the visual input.
\emph{Filler} spans receive lower weights, including hedging phrases (``maybe'', ``probably''), transitional phrases (``So then'', ``Now let me''), and verbatim repetition of the question.
Even when the final answer is correct, uncertain or poorly grounded reasoning receives reduced gradient, preventing accidental success from reinforcing undesirable reasoning patterns (Appendix~\S\ref{app:token_credit}). The per-token advantages $\hat{A}_t$ feed directly into the DAPO objective as token-level gradient weights.

\subsection{Policy Update}
\label{sec:dapo}
Mode-selection training requires maintaining sufficient exploration across all reasoning modes. DAPO~\citep{yu2025dapo} addresses this by preserving entropy during optimization while discarding rollout groups that provide no useful selection signal. DAPO extends GRPO~\cite{deepseek} with asymmetric clipping: a narrow lower bound $\varepsilon_{\mathrm{low}}$ tightly constrains probability-decreasing updates, while a wider upper bound $\varepsilon_{\mathrm{high}}$ preserves exploratory behavior. Let $N = \sum_{i,t}\mathbf{1}[q_{i,t}=1]$ denote the number of non-padding tokens. The per-token objective over the shaped advantages $\hat{A}_{i,t}$ is:
\begin{equation}
  \label{eq:dapo}
  \mathcal{J}_{\mathrm{DAPO}}(\theta)
  = \mathbb{E}\!\left[\frac{1}{N}
    \sum_{i,t} \mathbf{1}[q_{i,t}{=}1]\,
    \min\!\Bigl(r_{i,t}\hat{A}_{i,t},\,
      \mathrm{clip}(r_{i,t},1{-}\varepsilon_{\mathrm{low}},1{+}\varepsilon_{\mathrm{high}})\hat{A}_{i,t}
    \Bigr)\right],
\end{equation}
where $r_{i,t} = \pi_\theta(o_{i,t}\mid q,o_{i,<t})/\pi_{\theta_{\mathrm{old}}}(o_{i,t}\mid q,o_{i,<t})$ is the per-token probability ratio.

\section{Experiments}
\label{sec:experiments}
\paragraph{Implementation Details.}
We train two models: Qwen3-VL-8B~\cite{qwen3vl} and Qwen2.5-VL-7B~\cite{qwen2_5_vl}, both fine-tuned with LoRA~\cite{hu2022lora} using DeepSpeed ZeRO-3~\cite{rajbhandari2020zero} on 8 NVIDIA L40S and 4 A100 GPUs using the MS-Swift framework~\cite{zhao2025swift}.
We sample video frames at 1 fps, capped at 16 frames during training and 128 frames during evaluation (32 frames for Video-MMMU).
Training data for SFT and RL are described in Appendix~\S\ref{app:data_construction} and \S\ref{app:rl_setup}, and full hyperparameters are in Appendix~\S\ref{app:hyperparams}.
For evaluation, we prepend an adaptive selection prompt (Figure~\ref{fig:prompt_adaptive}) to each input. The model selects one of three reasoning modes (\tagDirect, \tagConcise, or \tagDeep) as its first generated tag. During training, we use a prioritized replay buffer to re-sample historically difficult inputs (Appendix~\S\ref{app:replay}).

\begin{table*}[!t]
\centering
\caption{\textbf{Comparison with state-of-the-art methods on image understanding and reasoning benchmarks.} Best scores per column \textbf{bolded} for each base model. Colored deltas show absolute change vs.\ the base model: {\color{darkgreen}gains} / {\color{darkred}losses}. Performance improvements of \modelname over the base model are shown in green with $95\%$ confidence interval (CI) margins ($\pm$) via paired bootstrap. $^{\ddagger}$Improvement not statistically significant at the $0.05$ level. All results are reproduced by us.}
\label{tab:sota_benchmarks}
\begin{adjustbox}{width=\textwidth,center}
\renewcommand{\arraystretch}{1.2}
\fontsize{7pt}{8pt}\selectfont
\setlength{\tabcolsep}{1mm}
\begin{tabular}{lcccccccc}
\toprule
& \multicolumn{7}{c}{\textbf{Image Understanding \& Reasoning Benchmarks}} \\
\cmidrule(lr){2-8}
\textbf{Model} &
\textbf{Avg Tokens $\downarrow$} &
\textbf{EMMA} & \textbf{MMMU} & \textbf{MathVista} &
\textbf{HallusionBench} & \textbf{AI2D} & \textbf{MM-Vet v2} \\
\midrule
\rowcolor{gray!7}
Qwen2.5-VL-7B  & -- & $25.8$ & $52.0$ & $68.4$ & $71.0$ & $82.7$ & $56.6$ \\
\quad$+$ SFT   & -- & $26.1$ \gain{0.3} & $52.9$ \gain{0.9} & $68.6$ \gain{0.2} & $70.5$ \loss{0.5} & $82.1$ \loss{0.6} & $56.7$ \gain{0.1} \\
\rowcolor{orange!10}
\multicolumn{8}{l}{\textit{Always Thinking}} \\
Video-R1~\cite{feng2025videor1}  & $342$ & $19.2$ \loss{6.6} & $52.5$ \gain{0.5} & $69.4$ \gain{1.0} & $64.4$ \loss{6.6} & $82.0$ \loss{0.7} & $55.8$ \loss{0.8} \\
Video-R2~\cite{maaz2025videor2}  & $411$ & $19.0$ \loss{6.8} & $49.6$ \loss{2.4} & $69.2$ \gain{0.8} & $63.6$ \loss{7.4} & $80.7$ \loss{2.0} & $56.0$ \loss{0.6} \\
VideoRFT~\cite{wang2025videorft} & $367$ & $19.5$ \loss{6.3} & $50.5$ \loss{1.5} & $69.1$ \gain{0.7} & $65.1$ \loss{5.9} & $81.3$ \loss{1.4} & $57.0$ \gain{0.4} \\
\rowcolor{orange!10}
\multicolumn{8}{l}{\textit{Auto Thinking}} \\
Video-Auto-R1~\cite{liu2026videoautor1} & $\mathbf{55}$ & $23.7$ \loss{2.1} & $52.0$ & $72.2$ \gain{3.8} & $66.2$ \loss{4.8} & $83.0$ \gain{0.3} & $55.4$ \loss{1.2} \\
\rowcolor{GroupColor1}
\textbf{\modelwithlogo} & $74$ & $\mathbf{27.3}$ \gainci{1.5}{0.5} & $\mathbf{54.5}$ \gainci{2.5}{0.7} & $\mathbf{73.8}$ \gainci{5.4}{0.6} & $\mathbf{71.8}$ \gainns{0.8}{0.9} & $\mathbf{83.2}$ \gainci{0.5}{0.3} & $\mathbf{57.9}$ \gainci{1.3}{0.8} \\
\midrule
\rowcolor{gray!7}
Qwen3-VL-8B  & -- & $22.4$ & $56.2$ & $74.1$ & $71.8$ & $80.3$ & $63.9$ \\
\quad$+$ SFT  & -- & $26.9$ \gain{4.5} & $56.2$ & $73.4$ \loss{0.7} & $71.4$ \loss{0.4} & $81.3$ \gain{1.0} & $64.1$ \gain{0.2} \\
\rowcolor{orange!10}
\multicolumn{8}{l}{\textit{Always Thinking}} \\
OneThinker~\cite{feng2025onethinker} & $539$ & $10.9$ \loss{11.5} & $61.2$ \gain{5.0} & $75.0$ \gain{0.9} & $54.3$ \loss{17.5} & $82.8$ \gain{2.5} & $64.7$ \gain{0.8} \\
\rowcolor{orange!10}
\multicolumn{8}{l}{\textit{Auto Thinking}} \\
Video-Auto-R1~\cite{liu2026videoautor1} & $259$ & $24.3$ \gain{1.9} & $\mathbf{61.7}$ \gain{5.5} & $72.7$ \loss{1.4} & $65.0$ \loss{6.8} & $83.9$ \gain{3.6} & $64.4$ \gain{0.5} \\
\rowcolor{GroupColor1}
\textbf{\modelwithlogo} & $\mathbf{108}$ & $\mathbf{30.8}$ \gainci{8.4}{0.5} & $61.5$ \gainci{5.3}{0.6} & $\mathbf{76.3}$ \gainci{2.2}{0.4} & $\mathbf{72.8}$ \gainns{1.0}{1.1} & $\mathbf{85.5}$ \gainci{5.2}{0.3} & $\mathbf{66.9}$ \gainci{3.0}{0.7} \\
\bottomrule
\end{tabular}
\end{adjustbox}
\end{table*}

\paragraph{Benchmarks.}
We evaluate on six image benchmarks (\textbf{EMMA}~\cite{hao2025emma}, \textbf{MMMU}~\cite{yue2024mmmu}, \textbf{MathVista}~\cite{lu2024mathvista}, \textbf{HallusionBench}~\cite{guan2024hallusionbench}, \textbf{AI2D}~\cite{kembhavi2016ai2d}, \textbf{MM-Vet~v2}~\cite{yu2024mmvetv2}) and five video benchmarks (\textbf{Video-Holmes}~\cite{videoholmes}, \textbf{Video-TT}~\cite{zhang2025videott}, \textbf{VSI-Bench}~\cite{yang2025vsibench}, \textbf{SciVideoBench}~\cite{deng2025scivideobench}, \textbf{Video-MMMU}~\cite{hu2025videommmu}).
Together, they span visual retrieval, hallucination robustness, mathematical and scientific reasoning, spatial understanding, and video-based knowledge acquisition.
Full descriptions are provided  in~\S\ref{app:eval_details}.

\paragraph{Baselines.}
We compare against two classes of reasoning models.
\textit{Always-thinking} models apply a fixed chain-of-thought strategy to every query. We evaluate Qwen2.5-VL-based methods including Video-R1~\cite{feng2025videor1}, Video-R2~\cite{maaz2025videor2}, and VideoRFT~\cite{wang2025videorft}, as well as the Qwen3-VL-based OneThinker~\cite{feng2025onethinker}.
\textit{Auto-thinking} models learn a binary decision between no reasoning and chain-of-thought reasoning, but use a single reasoning style when reasoning is invoked. We evaluate Video-Auto-R1~\cite{liu2026videoautor1} under both base models.
All baselines are reproduced under identical evaluation settings.

\subsection{Results}
\paragraph{Image benchmarks (Table~\ref{tab:sota_benchmarks}).}

On \textbf{Qwen2.5-VL}, \modelname achieves the best performance across all six image benchmarks while using only $74$ tokens on average. Always-thinking baselines perform poorly on perception-heavy tasks: all three models drop by an average of $6.6$ points on EMMA and HallusionBench, as long reasoning chains introduce plausible but visually ungrounded intermediate steps. Video-Auto-R1 reduces token usage to $55$ by learning a binary decision between no reasoning and chain-of-thought reasoning. However, its fixed decision threshold cannot distinguish between tasks that require visual retrieval and those that require reasoning, resulting in performance drops of $4.8$ points on HallusionBench and $1.2$ points on MM-Vet~v2.
On \textbf{Qwen3-VL}, \modelname leads on five of six image benchmarks, achieving $30.8$ on EMMA ($+8.4$ over the base model), $72.8$ on HallusionBench ($+1.0$), and $85.5$ on AI2D ($+5.2$), with an average of $108$ tokens. OneThinker, which applies long reasoning chains to all queries, suffers substantial drops of $11.5$ points on EMMA and $17.5$ points on HallusionBench.

\begin{table*}[!t]
\centering
\caption{\textbf{Comparison with state-of-the-art methods on video reasoning benchmarks.} Best scores per column \textbf{bolded} for each base model. Colored deltas show absolute change vs.\ the base model: {\color{darkgreen}gains} / {\color{darkred}losses}. Performance improvements of \modelname over the base model are shown in green with $95\%$ confidence interval (CI) margins ($\pm$) via paired bootstrap. All results are reproduced by us.}
\label{tab:sota_video_benchmarks}
\begin{adjustbox}{width=\textwidth,center}
\renewcommand{\arraystretch}{1.2}
\fontsize{7pt}{8pt}\selectfont
\setlength{\tabcolsep}{1mm}
\begin{tabular}{lcccccc}
\toprule
& \multicolumn{6}{c}{\textbf{Video Reasoning Benchmarks}} \\
\cmidrule(lr){2-7}
\textbf{Model} &
\textbf{Avg Tokens $\downarrow$} &
\textbf{Video-Holmes} & \textbf{Video-TT} & \textbf{VSI-Bench} &
\textbf{SciVideoBench} & \textbf{Video-MMMU} \\
\midrule
\rowcolor{gray!7}
Qwen2.5-VL-7B & -- & $43.2$ & $39.3$ & $31.8$ & $23.8$ & $52.1$ \\
\quad$+$ SFT  & -- & $44.7$ \gain{1.5} & $39.3$ & $33.0$ \gain{1.2} & $26.2$ \gain{2.4} & $53.3$ \gain{1.2} \\
\rowcolor{orange!10}
\multicolumn{7}{l}{\textit{Always Thinking}} \\
Video-R1~\cite{feng2025videor1}  & $360$ & $41.8$ \loss{1.4} & $40.1$ \gain{0.8} & $34.9$ \gain{3.1} & $27.0$ \gain{3.2} & $50.3$ \loss{1.8} \\
Video-R2~\cite{maaz2025videor2}  & $525$ & $40.7$ \loss{2.5} & $39.3$ & $31.9$ \gain{0.1} & $27.8$ \gain{4.0} & $48.3$ \loss{3.8} \\
VideoRFT~\cite{wang2025videorft} & $374$ & $42.9$ \loss{0.3} & $39.0$ \loss{0.3} & $\mathbf{36.0}$ \gain{4.2} & $28.0$ \gain{4.2} & $48.2$ \loss{3.9} \\
\rowcolor{orange!10}
\multicolumn{7}{l}{\textit{Auto Thinking}} \\
Video-Auto-R1~\cite{liu2026videoautor1} & $94$ & $47.7$ \gain{4.5} & $39.1$ \loss{0.2} & $35.2$ \gain{3.4} & $\mathbf{31.6}$ \gain{7.8} & $54.2$ \gain{2.1} \\
\rowcolor{GroupColor1}
\textbf{\modelwithlogo} & $\mathbf{59}$ & $\mathbf{48.0}$ \gainci{4.8}{0.6} & $\mathbf{41.0}$ \gainci{1.7}{0.4} & $35.8$ \gainci{4.0}{0.3} & $29.4$ \gainci{5.6}{0.7} & $\mathbf{56.9}$ \gainci{4.8}{0.8} \\
\midrule
\rowcolor{gray!7}
Qwen3-VL-8B & -- & $46.8$ & $39.7$ & $56.8$ & $29.4$ & $58.1$ \\
\quad$+$ SFT & -- & $47.9$ \gain{1.1} & $40.8$ \gain{1.1} & $56.8$ & $31.0$ \gain{1.6} & $58.5$ \gain{0.4} \\
\rowcolor{orange!10}
\multicolumn{7}{l}{\textit{Always Thinking}} \\
OneThinker~\cite{feng2025onethinker} & $421$ & $48.2$ \gain{1.4} & $40.0$ \gain{0.3} & $49.9$ \loss{6.9} & $32.6$ \gain{3.2} & $\mathbf{61.0}$ \gain{2.9} \\
\rowcolor{orange!10}
\multicolumn{7}{l}{\textit{Auto Thinking}} \\
Video-Auto-R1~\cite{liu2026videoautor1} & $284$ & $49.2$ \gain{2.4} & $40.2$ \gain{0.5} & $57.2$ \gain{0.4} & $32.0$ \gain{2.6} & $60.7$ \gain{2.6} \\
\rowcolor{GroupColor1}
\textbf{\modelwithlogo} & $\mathbf{137}$ & $\mathbf{49.9}$ \gainci{3.1}{0.5} & $\mathbf{41.2}$ \gainci{1.5}{0.3} & $\mathbf{58.2}$ \gainci{1.4}{0.2} & $\mathbf{33.1}$ \gainci{3.7}{0.6} & $60.5$ \gainci{2.4}{0.7} \\
\bottomrule
\end{tabular}
\end{adjustbox}

\end{table*}

\paragraph{Video benchmarks (Table~\ref{tab:sota_video_benchmarks}).}
On \textbf{Qwen2.5-VL}, \modelname achieves $48.0$ on Video-Holmes ($+4.8$ over the base model), $35.8$ on VSI-Bench ($+4.0$), and $56.9$ on Video-MMMU ($+4.8$) while using only $59$ tokens on average. Video-Holmes requires linking visual clues distributed across the entire video. Always-thinking baselines tend to expand early observations into long reasoning chains and prematurely commit to a hypothesis, causing all three methods to underperform the base model. On \textbf{Qwen3-VL}, \modelname achieves the best performance on four of five video benchmarks at $137$ tokens, including $49.9$ on Video-Holmes ($+3.1$ over the $46.8$ base) and $33.1$ on SciVideoBench ($+3.7$). OneThinker drops $6.9$ points on VSI-Bench, where spatial distances must be directly read from visual signals rather than inferred through extended reasoning chains. Across both base models, improvements are statistically significant in 20 of 22 comparisons under a paired bootstrap test. The only exception is HallusionBench, where both base models already score at least $71\%$.

\paragraph{RL gains over SFT.}
RL provides consistent improvements over the SFT initialization across all benchmarks. MMMU ($+5.3$), AI2D ($+4.2$), and EMMA ($+3.9$) on Qwen3-VL benefit the most, as all three require reasoning jointly over image and text and demand different reasoning modes depending on the query. SFT assigns modes based on fixed dataset heuristics and cannot adapt when queries fall outside these patterns. HallusionBench ($+1.4$) and Video-Holmes ($+2.0$) also improve due to the grounding reward and verifier feedback, which penalize ungrounded reasoning steps that SFT cannot suppress. SFT establishes mode-aware behavior but cannot determine which mode a given query requires. Rollout-based comparisons during RL enable the model to learn adaptive reasoning. RL also improves mode selection. For each training query, the controlled slots identify the utility-optimal mode $m^{*}$ (Eq.~\ref{eq:utility}), and we measure how often the adaptive slots select the same mode. Agreement increases from $40\%$ at the start of RL (the SFT checkpoint) to $75\%$ by the end of training. Thus, SFT teaches the policy the mode format, while RL teaches it which mode each query requires.

\begin{figure*}[!t]
  \centering
  \begin{minipage}[b]{0.32\linewidth}
    \centering
    \includegraphics[width=\linewidth]{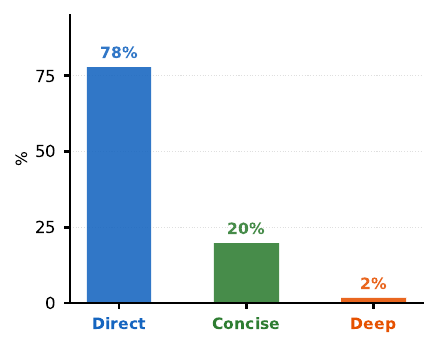}
    \par\vspace{2pt}{\small AI2D}
  \end{minipage}
  \hfill
  \begin{minipage}[b]{0.32\linewidth}
    \centering
    \includegraphics[width=\linewidth]{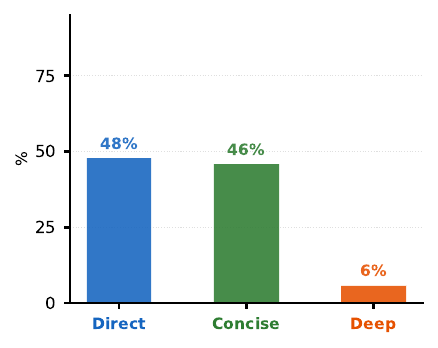}
    \par\vspace{2pt}{\small MMMU}
  \end{minipage}
  \hfill
  \begin{minipage}[b]{0.32\linewidth}
    \centering
    \includegraphics[width=\linewidth]{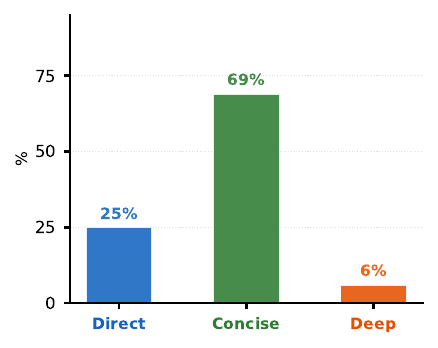}
    \par\vspace{2pt}{\small Video-Holmes}
  \end{minipage}
  \caption{\textbf{Reasoning mode distribution across representative benchmarks.} \modelname selects {\color{directcolor}\texttt{<}\textsc{direct}\texttt{>}} for simple visual retrieval (AI2D), shifts toward {\color{concisecolor}\texttt{<}\textsc{concise}\texttt{>}} as reasoning chains become necessary (MMMU, Video-Holmes), with {\color{deepcolor}\texttt{<}\textsc{deep}\texttt{>}} reserved for the small fraction of queries requiring hypothesis elimination. Full distributions across remaining benchmarks are in \S\ref{app:mode_dist}.}
  \label{fig:mode_dist_main}
\end{figure*}

\paragraph{Mode distribution.}
Figure~\ref{fig:mode_dist_main} shows the distribution of reasoning modes across three representative benchmarks. On AI2D, answers are read from a single diagram element, so the policy mostly selects \tagDirect. In contrast, MMMU requires combining visual evidence across heterogeneous image types with domain knowledge, leading to increased use of \tagConcise. Video-Holmes involves linking clues across video segments, resulting in a similar preference for \tagConcise, with selective use of \tagDeep for causal questions.

\subsection{Ablation Studies}
\label{sec:ablations}

\begin{table*}[t]
\centering
\footnotesize
\setlength{\tabcolsep}{3pt}
\begin{minipage}[t]{0.47\textwidth}
\centering
\caption{Each row ablates or replaces one training component while keeping all others fixed.}
\label{tab:ablation_components}
\resizebox{\linewidth}{!}{%
\begin{tabular}{l|cccc}
\toprule
\textbf{Variant} & \textbf{MMMU} & \textbf{EMMA} & \textbf{V-MMMU} & \textbf{Tok}$\downarrow$ \\
\midrule
$R_{\mathrm{ans}}$ + $R_{\mathrm{format}}$ only  & $51.8$ & $21.9$ & $49.8$ & $412$ \\
Only adaptive (no controlled slots)                 & $52.9$ & $24.6$ & $53.2$ & $198$ \\
Only controlled (no adaptive slots)                 & $53.6$ & $25.9$ & $53.0$ & $237$ \\
w/o $R_{\mathrm{select}}$                             & $53.1$ & $24.9$ & $54.3$ & $171$ \\
w/o $R_{\mathrm{verifier}}$                             & $54.1$ & $25.7$ & $55.5$ & $70$ \\
\midrule
\rowcolor{GroupColor1}
\textbf{\modelname} (full)                      & $\mathbf{54.5}$ & $\mathbf{27.3}$ & $\mathbf{56.9}$ & $\mathbf{67}$ \\
\bottomrule
\end{tabular}}
\end{minipage}
\hfill
\begin{minipage}[t]{0.47\textwidth}
\centering
\caption{Binary merges \tagConcise and \tagDeep into one thinking mode. Remaining rows ablate individual components.}
\label{tab:ablation_arch}
\resizebox{\linewidth}{!}{%
\begin{tabular}{l|cccc}
\toprule
\textbf{Variant} & \textbf{MMMU} & \textbf{EMMA} & \textbf{V-MMMU} & \textbf{Tok}$\downarrow$ \\
\midrule
Binary (2-mode: direct / think)        & $52.9$ & $24.4$ & $55.1$ & $\mathbf{61}$ \\
No token credit map                    & $54.2$ & $26.0$ & $55.8$ & $69$ \\
Frozen verifier (no DPO refresh)       & $53.8$ & $26.1$ & $55.3$ & $68$ \\
Utility w/o cost factor                & $54.3$ & $27.1$ & $56.6$ & $394$ \\
\midrule
\rowcolor{GroupColor1}
\textbf{\modelname} (full)                      & $\mathbf{54.5}$ & $\mathbf{27.3}$ & $\mathbf{56.9}$ & $67$ \\
\bottomrule
\end{tabular}}
\end{minipage}

\end{table*}

\paragraph{RQ1: Does the rollout structure matter?}
Yes, all three slot types are necessary (Table~\ref{tab:ablation_components}).  RL (row 1), which trains with only $R_{\mathrm{ans}}$ and $R_{\mathrm{format}}$, produces the lowest accuracy and highest token usage ($412$), showing that without mode-structured rollouts, the policy defaults to a single undifferentiated reasoning strategy and fails to learn cost-aware behavior. Removing the \textit{controlled slots} (row 2) eliminates within-group comparisons. As a result, EMMA drops to $24.6$, since the adaptive slots no longer receive supervision about how alternative reasoning modes would have performed on the same query. Removing the \textit{adaptive slots} (row 3) eliminates mode selection during training. Without a mode-selection objective, \tagDirect is too risky (low accuracy), and \tagDeep is suppressed by cost pressure, so the policy settles on \tagConcise, increasing token usage to $237$.

\paragraph{RQ2: Do the reward signals supervising adaptive slots contribute?}
Yes, both reward components are necessary (Table~\ref{tab:ablation_components}, rows 4--5). Dropping $R_{\mathrm{select}}$ (row 4) removes supervision from adaptive slots; without it, the policy cannot identify which mode is most effective per query, causing accuracy to drop (EMMA $24.9$) while token usage rises to $171$ as mode selection becomes unguided. Dropping $R_{\mathrm{verifier}}$ (row 5) leaves MMMU largely unchanged ($54.1$) but reduces EMMA to $25.7$. This confirms that the verifier primarily improves grounding quality rather than overall reasoning accuracy.

\paragraph{RQ3: Does the three-mode space outperform a binary switch?} Yes, three reasoning modes outperform a binary setup (Table~\ref{tab:ablation_arch}, row 1). Collapsing \tagConcise and \tagDeep into a single thinking mode reduces EMMA to $24.4$ and Video-MMMU to $55.1$. The low token count ($61$) reflects cost-driven reliance on \tagDirect, as it avoids reasoning for medium-difficulty queries. Furthermore, performance becomes comparable to Video-Auto-R1 ($23.7$, $54.2$), showing that distinguishing between compositional and deliberative reasoning is necessary for adaptive reasoning. The \tagConcise and \tagDeep boundary is also stable under a mode swap (Appendix Table~\ref{tab:mode_swap}). Forcing \tagConcise on queries assigned to \tagDeep lowers accuracy by $7$ to $9$ points, while forcing \tagDeep on queries assigned to \tagConcise helps EMMA ($+4$) at $8.8\times$ the tokens but hurts Video-Holmes ($-6$).

\paragraph{RQ4: How does each grounding component contribute?} All three components contribute (Table~\ref{tab:ablation_arch}, rows 2--4). Removing the token-level credit map (row 2) reduces EMMA to $26.0$ while leaving MMMU largely unchanged, indicating that credit shaping improves grounding quality without substantially affecting predicted answers. Freezing the verifier without DPO refresh (row 3) causes gradual degradation on grounding-sensitive benchmarks (EMMA $26.1$, Video-MMMU $55.3$), suggesting that policy improvements eventually outpace verifier calibration. Because the verifier is refreshed on policy rollouts, the policy could learn to earn high verifier scores without answering better. EMMA accuracy is scored against ground-truth answers rather than by the verifier, and it rises from training without $R_{\mathrm{verifier}}$ ($25.7$) to a frozen verifier ($26.1$) to the DPO-refreshed verifier ($27.3$). The refresh therefore improves the answers themselves, not only the verifier scores. Removing the cost factor from the utility function (row 4) leaves accuracy nearly unchanged but increases token usage to $394$, as the policy defaults to \tagConcise or \tagDeep even for easy queries.

\begin{table*}[t]
\centering
\footnotesize
\setlength{\tabcolsep}{3pt}
\begin{minipage}[t]{0.53\textwidth}
\centering
\caption{Simpler alternatives to \modelname on Qwen2.5-VL-7B.}
\label{tab:simple_baselines}
\resizebox{\linewidth}{!}{%
\begin{tabular}{l|ccccc}
\toprule
\textbf{Method} & \textbf{MMMU} & \textbf{EMMA} & \textbf{V-Holmes} & \textbf{V-MMMU} & \textbf{Tok}$\downarrow$ \\
\midrule
Qwen2.5-VL-7B                              & $52.0$ & $25.8$ & $43.2$ & $52.1$ & -- \\
Zero-shot prompt selector                  & $51.8$ & $25.2$ & $42.8$ & $51.5$ & $340$ \\
Static budget ($64$ tokens)                & $44.5$ & $18.2$ & $33.8$ & $42.0$ & $64$ \\
SFT $+$ Lightweight router                  & $53.5$ & $26.5$ & $46.0$ & $54.5$ & $85$ \\
\midrule
\rowcolor{GroupColor1}
\textbf{\modelname}                  & $\mathbf{54.5}$ & $\mathbf{27.3}$ & $\mathbf{48.0}$ & $\mathbf{56.9}$ & $\mathbf{61}$ \\
\bottomrule
\end{tabular}}
\end{minipage}
\hfill
\begin{minipage}[t]{0.42\textwidth}
\centering
\caption{Reasoning mode forced or left adaptive at inference.}
\label{tab:ablation_forced}
\resizebox{\linewidth}{!}{%
\begin{tabular}{l|cccc}
\toprule
\textbf{Mode} & \textbf{MMMU} & \textbf{EMMA} & \textbf{V-MMMU} & \textbf{Tok}$\downarrow$ \\
\midrule
Forced \tagDirect  & 51.4 & 22.1 & 49.7 & \textbf{57} \\
Forced \tagConcise & 53.1 & 25.6 & 53.2 & 247 \\
Forced \tagDeep    & 52.8 & \textbf{28.5} & 55.8 & 963 \\
\midrule
\rowcolor{GroupColor1}
\textbf{\modelname}   & \textbf{54.5} & 27.3 & \textbf{56.9} & 67 \\
\bottomrule
\end{tabular}}
\end{minipage}
\end{table*}

\paragraph{RQ5: Do simpler alternatives achieve comparable performance?}
No simpler alternative matches \modelname (Table~\ref{tab:simple_baselines}). All three baselines build on Qwen2.5-VL-7B. The zero-shot prompt selector prepends the adaptive prompt (Figure~\ref{fig:prompt_adaptive}) to the untrained base model. It provides no gain over the base model and uses $340$ tokens on average, so prompting alone does not produce cost-aware mode selection. The static budget caps the output of the base model at $64$ tokens, close to the $61$ tokens of \modelname. This cap reduces performance across all four benchmarks, including EMMA from $25.8$ to $18.2$, because reasoning is cut off mid-chain. Thus, shorter responses alone do not explain the gains. The lightweight router is a linear classification head on the last hidden state of the frozen base model, trained on utility-optimal mode labels. It routes each query to the SFT model with the corresponding mode prompt (Appendix Figures~\ref{fig:prompt_direct}--\ref{fig:prompt_deep}), isolating mode selection from generation quality. The router outperforms the base model on all four benchmarks but still trails \modelname by $2.0$ points on Video-Holmes and $2.4$ points on Video-MMMU, indicating that RL improves both routing and generation within each mode.

\paragraph{RQ6: Can the right fixed mode match adaptive selection?}
No fixed reasoning mode Pareto-dominates adaptive selection (Table~\ref{tab:ablation_forced}). \tagDirect achieves the lowest token usage but performs poorly on reasoning-heavy benchmarks, dropping from $27.3$ to $22.1$ on EMMA. In contrast, \tagDeep improves EMMA performance but incurs substantial token overhead on image tasks where extended reasoning is unnecessary, and slightly underperforms \tagConcise on MMMU. No fixed mode achieves \tagAdaptive's combination of accuracy and efficiency. The $1.2$-point EMMA gap to forced \tagDeep is a deliberate trade-off (Appendix Table~\ref{tab:emma_partition}). Forced \tagDeep gains $4$ points only on queries routed to \tagConcise, at $8.8\times$ the tokens, and loses $2$ points on queries routed to \tagDirect.

\section{Conclusion}

We present \modelname, a training framework that enables multimodal LLMs to adapt their reasoning mode to each query without per-query annotations. We show that adaptive reasoning can be learned through structured rollout comparisons across \textsc{Percept}, \textsc{Compose}, and \textsc{Deliberate} modes. Our experiments show that uniform reasoning fails in two complementary ways, as over-reasoning on perceptual tasks introduces hallucinated intermediate steps and insufficient deliberation on ambiguous tasks leads to incorrect interpretations. By adapting how to reason rather than only whether to reason, \modelname addresses both failure modes simultaneously and consistently improves accuracy on image and video benchmarks while substantially reducing token usage compared to always-thinking strategies. Future work includes tool-augmented modes, mid-response mode switching when the initial mode proves insufficient, early exit once the evidence suffices, adaptive reasoning across other modalities, and reasoning in latent space rather than in text.

\section*{Acknowledgments}
We thank Pritam Sarkar for helpful discussions and feedback throughout this work. This research was supported by the National Eye Institute (NEI) of the National Institutes of Health (NIH) under award number R01EY034562.\ The content is solely the responsibility of the authors and does not necessarily represent the official views of the NIH. We acknowledge Research Computing at Arizona State University for providing computing resources.

\bibliographystyle{abbrvnat}
\bibliography{references}

\medskip

\appendix

\newpage
\section*{Appendix}
\label{app:appendix_root}

\addcontentsline{toc}{section}{Appendix Table of Contents}
\startcontents[appendix]
\printcontents[appendix]{l}{1}{\setcounter{tocdepth}{2}}

\clearpage
\section{Related Work}
\paragraph{Multimodal grounding and video reasoning.}
Video understanding demands temporal localization, spatial grounding, and event-level reasoning~\citep{videomme,videoholmes,mmvu,tomato,heilbron2015activitynet,gao2017tall}, yet these demands vary sharply across queries.
Work on the perception-reasoning gap shows that token-level balance between visual extraction and chain-of-thought elaboration is critical~\citep{lu2026bridging}, and that pixel-level motion grounding~\citep{deng2025motion}, targeted supervision~\citep{lee2026visioncoach,zhong2025rethinking}, and structured evidence use~\citep{zhang2025thinking,rasheed2025videocom,cheng2026graphthinker} each improve grounding in different ways.
Work on multimodal reasoning structure~\citep{chen2025bring,chung2025mllms,li2025perception,chen2025perception} establishes that perception and reasoning are distinct cognitive operations with different informational requirements, and that applying a single generation style across both leads to systematic errors.
This motivates \modelname's three-mode design: rather than a single chain-of-thought, the model selects the reasoning structure whose demands match the query, with grounding signals integrated directly into the reward and token-level credit.
\begin{figure*}[t]
  \centering
  \begin{minipage}[b]{0.48\linewidth}
    \centering
    \includegraphics[width=\linewidth]{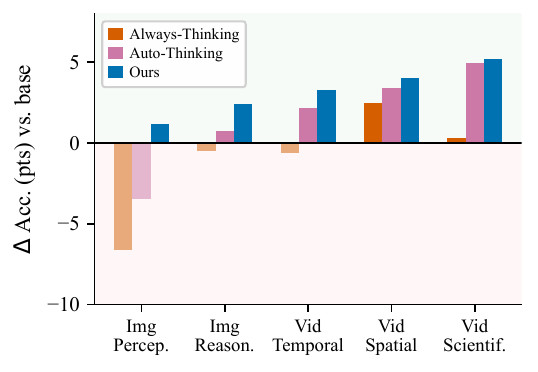}
    \par\vspace{2pt}{\small Qwen2.5-VL}
  \end{minipage}
  \hfill
  \begin{minipage}[b]{0.48\linewidth}
    \centering
    \includegraphics[width=\linewidth]{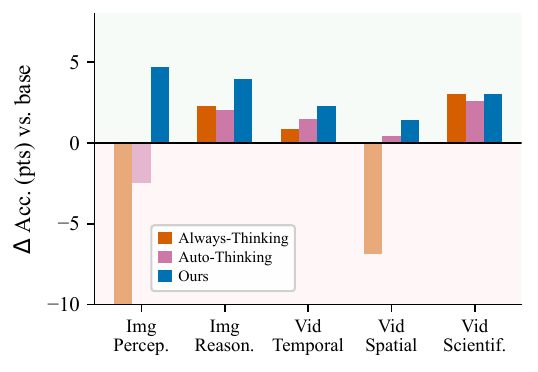}
    \par\vspace{2pt}{\small Qwen3-VL}
  \end{minipage}
  \caption{\textbf{Per-category gains over the base model.}
    Each bar shows the absolute accuracy change vs.\ the base model across five task categories (averaged over benchmarks within each group).
    Always-thinking models collapse on image perception (hallucination pressure) and, on Qwen3-VL, on spatial reasoning. Auto-thinking reduces the perception loss but still falls below the base model there.
    \modelname is the only method with positive gains across all five categories in both base models.}
  \label{fig:analysis_plots}
\end{figure*}

\section{Ablation Studies}
\label{app:additional_ablations}
\label{app:reward_ablation}

\paragraph{RQ7: Do per-category gains hold across model families?}

Figure~\ref{fig:analysis_plots} shows per-category gains over the base model.
Five categories aggregate benchmarks as follows: \textit{Image Perception} (EMMA, HallusionBench), \textit{Image Reasoning} (MMMU, MathVista, AI2D, MM-Vet~v2), \textit{Video Temporal} (Video-Holmes, Video-TT), \textit{Video Spatial} (VSI-Bench), and \textit{Video Scientific} (SciVideoBench, Video-MMMU).
Always-thinking collapses on Image Perception (Qwen2.5-VL: $-6.6$ pts avg, Qwen3-VL: $-14.5$ pts avg) and Video Spatial ($-6.9$ pts on Qwen3-VL) because verbose thinking hallucinates answers that must be read from the visual signal.
Auto-thinking gains in the other four categories but still loses on Image Perception ($-3.5$ pts avg on Qwen2.5-VL and $-2.5$ on Qwen3-VL), because its binary decision between no reasoning and chain-of-thought cannot distinguish queries that require visual retrieval from those that require reasoning.
\modelname is the only method with positive gains across all five categories in both families.
Qwen3-VL shows larger absolute gains on Image Perception ($+4.7$ pts) and Image Reasoning ($+3.9$ pts), reflecting a stronger base model that benefits more from accurate mode selection on hard reasoning tasks.
\begin{figure*}[t]
  \centering
  \begin{minipage}[b]{0.24\linewidth}
    \centering
    \includegraphics[width=\linewidth]{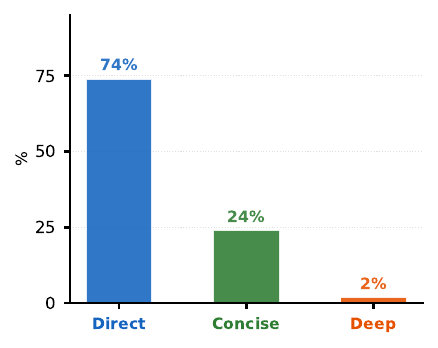}
    \par\vspace{2pt}{\small HallusionBench}
  \end{minipage}
  \hfill
  \begin{minipage}[b]{0.24\linewidth}
    \centering
    \includegraphics[width=\linewidth]{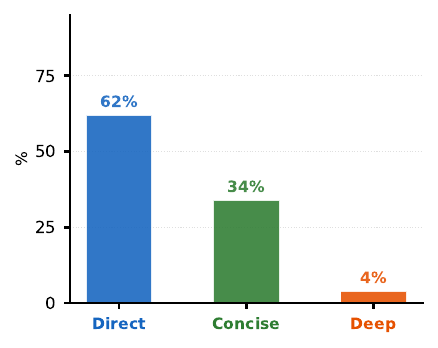}
    \par\vspace{2pt}{\small MM-Vet v2}
  \end{minipage}
  \hfill
  \begin{minipage}[b]{0.24\linewidth}
    \centering
    \includegraphics[width=\linewidth]{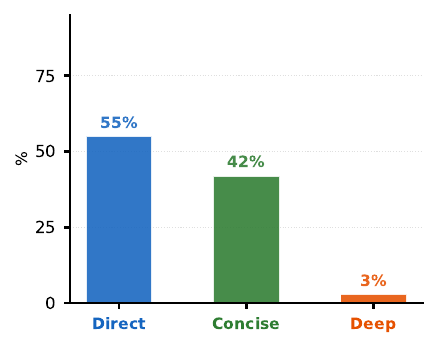}
    \par\vspace{2pt}{\small\textit{Video-TT}}
  \end{minipage}
  \hfill
  \begin{minipage}[b]{0.24\linewidth}
    \centering
    \includegraphics[width=\linewidth]{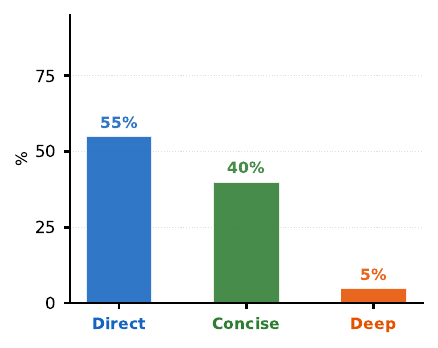}
    \par\vspace{2pt}{\small MathVista}
  \end{minipage}

  \vspace{0.3cm}
  \begin{minipage}[b]{0.24\linewidth}
    \centering
    \includegraphics[width=\linewidth]{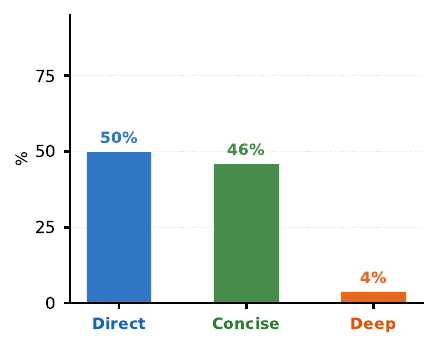}
    \par\vspace{2pt}{\small VSI-Bench}
  \end{minipage}
  \hfill
  \begin{minipage}[b]{0.24\linewidth}
    \centering
    \includegraphics[width=\linewidth]{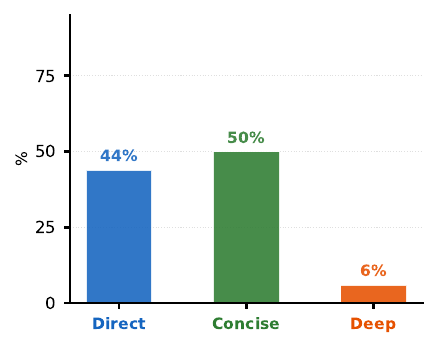}
    \par\vspace{2pt}{\small EMMA}
  \end{minipage}
  \hfill
  \begin{minipage}[b]{0.24\linewidth}
    \centering
    \includegraphics[width=\linewidth]{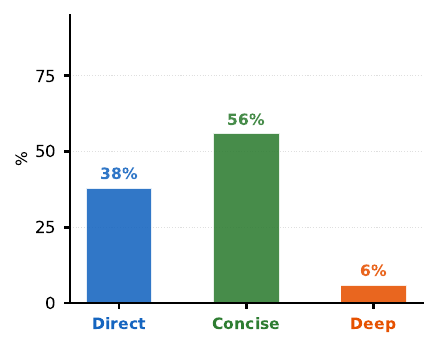}
    \par\vspace{2pt}{\small Video-MMMU}
  \end{minipage}
  \hfill
  \begin{minipage}[b]{0.24\linewidth}
    \centering
    \includegraphics[width=\linewidth]{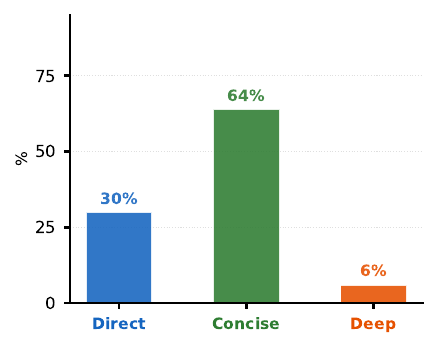}
    \par\vspace{2pt}{\small SciVideoBench}
  \end{minipage}
  \caption{\textbf{Reasoning mode distribution across remaining benchmarks.} The {\color{directcolor}\texttt{<}\textsc{direct}\texttt{>}}-to-{\color{concisecolor}\texttt{<}\textsc{concise}\texttt{>}} shift tracks benchmark cognitive demand: retrieval-oriented tasks (HallusionBench, Video-TT) remain {\color{directcolor}\texttt{<}\textsc{direct}\texttt{>}}-heavy, while inference-heavy and research-level benchmarks (EMMA, SciVideoBench) shift toward {\color{concisecolor}\texttt{<}\textsc{concise}\texttt{>}}. {\color{deepcolor}\texttt{<}\textsc{deep}\texttt{>}} usage stays at most 6\% throughout.}
  \label{fig:mode_dist_appendix}
\end{figure*}

\paragraph{RQ8: Does mode selection reflect query cognitive demand?}
\label{app:mode_dist}
Figure~\ref{fig:mode_dist_appendix} shows the mode distribution across the remaining 8 benchmarks.
\tagDirect usage spans $74\%$ (HallusionBench) to $30\%$ (SciVideoBench), a $44$-point range that tracks each benchmark's cognitive profile rather than cost alone. A cost-only heuristic would maximize \tagDirect everywhere.
HallusionBench ($74\%$ \tagDirect) is designed so that correct answers require reading the image directly rather than reasoning from priors~\citep{guan2024hallusionbench}; extended chains bypass the visual signal rather than grounding in it.
EMMA ($50\%$ \tagConcise) is the only image benchmark where \tagConcise is the most frequent mode, as its questions require integrating visual observations with domain reasoning over multiple steps~\citep{hao2025emma}.
SciVideoBench ($64\%$ \tagConcise) draws from research-level experimental videos across 25+ scientific disciplines; answering requires chaining precise spatiotemporal observations with expert domain knowledge~\citep{deng2025scivideobench}.
\tagDeep stays at most $6\%$ across all benchmarks, rising only on the small fraction of questions where multiple competing explanations must be actively ruled out.

\begin{table*}[t]
\centering
\caption{
  \textbf{Reward component and mode-space ablations.}
  Rows R1 and R2 each remove one reward component and Row R3 removes grounding from the utility ($\delta{=}0$), while all other components remain active (\checkmark).
  In Row R4 all reward components remain active but \tagDeep is removed, restricting the model to \tagDirect and \tagConcise only.}

\setlength{\tabcolsep}{4pt}
\label{tab:reward_ablation}
\begin{adjustbox}{max width=\linewidth,center}
\renewcommand{\arraystretch}{1.15}
\begin{tabular}{l ccc | ccccc}
\toprule
\textbf{Configuration}
  & $R_{\mathrm{balance}}$ & $R_{\mathrm{cost}}$ & $\delta$
  & \textbf{EMMA} & \textbf{HallusionBench} & \textbf{Video-Holmes} & \textbf{VSI-Bench} & \textbf{SciVideoBench} \\
\midrule
Qwen2.5-VL-7B
  & -- & -- & --
  & $25.8$ & $71.0$ & $43.2$ & $31.8$ & $23.8$ \\
\midrule
(R1)~w/o mode diversity ($R_{\mathrm{balance}}$)
  & $\times$ & \checkmark & \checkmark
  & $25.3$ & $71.4$ & $44.9$ & $33.7$ & $26.3$ \\
(R2)~w/o length pressure ($R_{\mathrm{cost}}$)
  & \checkmark & $\times$ & \checkmark
  & $27.1$ & $69.4$ & $47.2$ & $33.1$ & $28.6$ \\
(R3)~$\delta{=}0$ (accuracy-only utility)
  & \checkmark & \checkmark & $\times$
  & $26.7$ & $70.6$ & $45.6$ & $32.9$ & $27.2$ \\
\midrule
(R4)~\tagDirect $+$ \tagConcise only (no \tagDeep)
  & \checkmark & \checkmark & \checkmark
  & $26.0$ & $71.3$ & $46.1$ & $34.5$ & $27.8$ \\
\midrule
\rowcolor{GroupColor1}
\modelname (full)
  & \checkmark & \checkmark & \checkmark
  & $\mathbf{27.3}$ & $\mathbf{71.8}$ & $\mathbf{48.0}$ & $\mathbf{35.8}$ & $\mathbf{29.4}$ \\
\bottomrule
\end{tabular}
\end{adjustbox}
\end{table*}

\begin{table*}[t]
\centering
\footnotesize
\setlength{\tabcolsep}{5pt}
\caption{%
  \textbf{Design decision sensitivity.}
  Each block varies one hyperparameter while holding all others at the chosen value (\colorbox{GroupColor1}{highlighted}).
}
\label{tab:sensitivity}
\begin{adjustbox}{max width=\linewidth,center}
\renewcommand{\arraystretch}{1.15}
\begin{tabular}{ll | ccc ccc}
\toprule
\textbf{Decision} & \textbf{Setting} & \textbf{EMMA} & \textbf{MMMU} & \textbf{HallusionBench} & \textbf{Video-Holmes} & \textbf{VSI-Bench} & \textbf{SciVideoBench} \\
\midrule
Qwen2.5-VL-7B & -- & $25.8$ & $52.0$ & $71.0$ & $43.2$ & $31.8$ & $23.8$ \\
\midrule
\multirow{2}{*}{Grounding weight $\delta$}
  & \cellcolor{GroupColor1}$\delta=0.35$ \textit{(ours)} & \cellcolor{GroupColor1}$\mathbf{27.3}$ & \cellcolor{GroupColor1}$\mathbf{54.5}$ & \cellcolor{GroupColor1}$\mathbf{71.8}$ & \cellcolor{GroupColor1}$\mathbf{48.0}$ & \cellcolor{GroupColor1}$35.8$ & \cellcolor{GroupColor1}$\mathbf{29.4}$ \\
  & $\delta=0.70$                                & $26.8$ & $53.1$ & $71.3$ & $47.4$ & $\mathbf{36.2}$ & $28.7$ \\
\midrule
\multirow{3}{*}{Selection weight $w_{\mathrm{select}}$}
  & $w_{\mathrm{select}}=0.20$  & $26.2$ & $54.1$ & $71.5$ & $46.5$ & $35.1$ & $28.3$ \\
  & \cellcolor{GroupColor1}$w_{\mathrm{select}}=0.40$ \textit{(ours)} & \cellcolor{GroupColor1}$\mathbf{27.3}$ & \cellcolor{GroupColor1}$\mathbf{54.5}$ & \cellcolor{GroupColor1}$\mathbf{71.8}$ & \cellcolor{GroupColor1}$\mathbf{48.0}$ & \cellcolor{GroupColor1}$\mathbf{35.8}$ & \cellcolor{GroupColor1}$\mathbf{29.4}$ \\
  & $w_{\mathrm{select}}=0.60$  & $26.9$ & $53.8$ & $71.4$ & $47.6$ & $35.4$ & $29.0$ \\
\bottomrule
\end{tabular}
\end{adjustbox}
\end{table*}

\paragraph{RQ9: Do the balance, cost, and grounding terms each contribute?} Table~\ref{tab:reward_ablation} ablates $R_{\mathrm{balance}}$, $R_{\mathrm{cost}}$, and the grounding weight $\delta$ one at a time.
Row R1 removes $R_{\mathrm{balance}}$: without the diversity reward, the model collapses toward \tagDirect on most queries. Video-Holmes drops to $44.9$ ($-3.1$) and SciVideoBench to $26.3$ ($-3.1$) since both require mode diversity across questions. HallusionBench is largely unaffected ($71.4$, $-0.4$) because \tagDirect is already the appropriate mode there.
Row R2 removes $R_{\mathrm{cost}}$: without length pressure, the model drifts toward verbose responses. HallusionBench drops to $69.4$ ($-2.4$) and VSI-Bench to $33.1$ ($-2.7$) because extended chains produce visually ungrounded steps on tasks where the answer must be read from the visual signal. EMMA is largely unaffected ($27.1$, $-0.2$) as reasoning tasks tolerate longer outputs.
Row R3 sets $\delta{=}0$, removing grounding from the utility function. VSI-Bench drops to $32.9$ ($-2.9$) and Video-Holmes to $45.6$ ($-2.4$) as both require accurate spatial and temporal grounding to select the right mode; without $\delta$ the utility cannot distinguish a grounded from an ungrounded correct answer.

\paragraph{RQ10: Is \tagDeep necessary or does \tagConcise cover its role?}
Row R4 of Table~\ref{tab:reward_ablation} removes \tagDeep entirely, restricting the model to \tagDirect and \tagConcise.
Video-Holmes drops to $46.1$ ($-1.9$) and SciVideoBench to $27.8$ ($-1.6$), as both contain queries that require ruling out competing hypotheses across evidence, which \tagConcise cannot fully substitute.
HallusionBench is flat ($71.3$, $-0.5$) since the benchmark is dominated by \tagDirect queries where \tagDeep is rarely selected.
The results confirm that \tagDeep contributes independently: its removal is not recoverable by \tagConcise, and the benchmarks most sensitive to this removal are precisely those with hypothesis-testing demand.

\paragraph{RQ11: How sensitive are the key hyperparameters?}
Table~\ref{tab:sensitivity} sweeps $\delta$ and $w_{\mathrm{select}}$ while holding all other parameters fixed.
For $\delta{=}0.70$, excess grounding emphasis biases utility toward modes that produce grounding tags regardless of accuracy; MMMU drops to $53.1$ ($-1.4$) and HallusionBench to $71.3$ ($-0.5$) as grounding overrides the accuracy signal on tasks where visual lookup suffices; VSI-Bench increases slightly ($+0.4$) since spatial grounding is genuinely informative there.
For $w_{\mathrm{select}}{=}0.20$, the weak selection signal allows cost pressure to suppress \tagDeep even when utility identifies it as optimal. Hence, Video-Holmes drops to $46.5$ ($-1.5$) and EMMA to $26.2$ ($-1.1$). This occurs because the adaptive slot's penalty for selecting a suboptimal mode is too small to outweigh the cost reward favoring shorter responses.
For $w_{\mathrm{select}}{=}0.60$, the stronger penalty starts to drown out verifier feedback, slightly degrading MMMU ($53.8$, $-0.7$) and HallusionBench ($71.4$, $-0.4$).
Both parameters are robust within a moderate range, all settings remain above the base model, and the chosen values ($\delta{=}0.35$, $w_{\mathrm{select}}{=}0.40$) achieve the best overall balance.

\begin{table*}[t]
\centering
\footnotesize
\setlength{\tabcolsep}{3pt}
\begin{minipage}[t]{0.46\textwidth}
\centering
\caption{Mode swap at inference. Each row forces the other mode on queries that \modelname assigns to the source mode.}
\label{tab:mode_swap}
\begin{adjustbox}{max width=\linewidth,center}
\begin{tabular}{ll|cccc}
\toprule
\textbf{Swap} & \textbf{Benchmark} & \textbf{Own} & \textbf{Swapped} & $\boldsymbol{\Delta}$ & \textbf{Tok} \\
\midrule
\tagDeep $\rightarrow$ \tagConcise & EMMA         & $33$ & $26$ & $-7$ & $480 \rightarrow 100$ \\
\tagDeep $\rightarrow$ \tagConcise & Video-Holmes & $55$ & $46$ & $-9$ & $500 \rightarrow 120$ \\
\midrule
\tagConcise $\rightarrow$ \tagDeep & EMMA         & $27$ & $31$ & $+4$ & $100 \rightarrow 880$ \\
\tagConcise $\rightarrow$ \tagDeep & Video-Holmes & $50$ & $44$ & $-6$ & $120 \rightarrow {>}1{,}000$ \\
\bottomrule
\end{tabular}
\end{adjustbox}
\end{minipage}
\hfill
\begin{minipage}[t]{0.50\textwidth}
\centering
\caption{EMMA partitioned by the mode that \modelname selects, compared with forced \tagDeep on the same queries. $\Delta$ is adaptive minus forced \tagDeep accuracy.}
\label{tab:emma_partition}
\begin{adjustbox}{max width=\linewidth,center}
\begin{tabular}{l|c|cc|cc|c}
\toprule
& & \multicolumn{2}{c|}{\textbf{Adaptive}} & \multicolumn{2}{c|}{\textbf{Forced} \tagDeep} & \\
\textbf{Selected mode} & \textbf{\#} & \textbf{Acc} & \textbf{Tok} & \textbf{Acc} & \textbf{Tok} & $\boldsymbol{\Delta}$ \\
\midrule
\tagDirect ($44\%$)  & $1{,}227$ & $27$ & $30$  & $25$ & $770$     & $+2$ \\
\tagConcise ($50\%$) & $1{,}394$ & $27$ & $100$ & $31$ & $880$     & $-4$ \\
\tagDeep ($6\%$)     & $167$     & $33$ & $480$ & $33$ & ${>}1{,}000$ & $0$ \\
\midrule
\rowcolor{GroupColor1}
All                  & $2{,}788$ & $27.3$ & $92$ & $28.5$ & $845$ & $-1.2$ \\
\bottomrule
\end{tabular}
\end{adjustbox}
\end{minipage}
\end{table*}

\paragraph{RQ12: Is the \tagConcise and \tagDeep boundary stable?}
Table~\ref{tab:mode_swap} swaps modes at inference on queries that \modelname assigns to \tagDeep or \tagConcise. For the $\approx$280 queries assigned to \tagDeep, forcing \tagConcise lowers accuracy by $7$ points on EMMA and $9$ points on Video-Holmes, so \tagConcise cannot recover what \tagDeep provides. Forcing \tagDeep on a matched set of \tagConcise queries is benchmark-dependent. EMMA gains $4$ points but uses $8.8\times$ more tokens, while Video-Holmes loses $6$ points because unnecessary hypothesis generation disrupts sequential reasoning. The boundary is therefore utility-aware rather than heuristic, and \tagDeep is invoked when its accuracy gain justifies its cost, which is the objective of the utility in Eq.~\ref{eq:utility}.

\paragraph{RQ13: Does the policy under-invoke \tagDeep on EMMA?}
Forced \tagDeep reaches $28.5$ on EMMA compared with $27.3$ for adaptive selection (Table~\ref{tab:ablation_forced}). Table~\ref{tab:emma_partition} partitions EMMA by the mode that \modelname selects and evaluates forced \tagDeep on the same queries. On queries routed to \tagDirect, adaptive selection is $2$ points more accurate because over-reasoning hurts perceptual queries. On queries routed to \tagConcise, forced \tagDeep gains $4$ points but uses about $880$ tokens instead of $100$. On queries routed to \tagDeep, both reach the same accuracy, and adaptive \tagDeep uses fewer tokens because $R_{\mathrm{cost}}$ teaches the model to reason sufficiently rather than exhaustively. The $1.2$-point gap is therefore a deliberate accuracy and cost trade-off, and forced \tagDeep also underperforms adaptive selection on MMMU and Video-MMMU (Table~\ref{tab:ablation_forced}).

\begin{table*}[!t]
\centering
\footnotesize
\setlength{\tabcolsep}{4pt}
\begin{minipage}[t]{0.44\textwidth}
\centering
\caption{General capability of the verifier before and after DPO refresh.}
\label{tab:verifier_drift}
\begin{adjustbox}{max width=\linewidth,center}
\renewcommand{\arraystretch}{1.15}
\begin{tabular}{l|cc|c}
\toprule
\textbf{Benchmark} & \textbf{Base Qwen3-VL-8B} & \textbf{DPO-updated} & $\boldsymbol{\Delta}$ \\
\midrule
MMMU           & $56.2$ & $56.0$ & $-0.2$ \\
HallusionBench & $71.8$ & $72.0$ & $+0.2$ \\
MathVista      & $74.1$ & $73.9$ & $-0.2$ \\
VSI-Bench      & $56.8$ & $57.1$ & $+0.3$ \\
\bottomrule
\end{tabular}
\end{adjustbox}
\end{minipage}
\hfill
\begin{minipage}[t]{0.52\textwidth}
\centering
\caption{Video-Holmes routing against an independent GPT-5.5 mode judgment. The last column forces the judged mode.}
\label{tab:holmes_routing}
\begin{adjustbox}{max width=\linewidth,center}
\renewcommand{\arraystretch}{1.15}
\begin{tabular}{l|c|cc}
\toprule
\textbf{Routing} & \textbf{\# Queries} & \textbf{Acc (adaptive)} & \textbf{Acc (judged mode)} \\
\midrule
Agrees with judge ($72\%$)    & $1{,}323$ & $51$ & $51$ \\
Disagrees with judge ($28\%$) & $514$     & $40$ & $47$ \\
\midrule
\rowcolor{GroupColor1}
Gap                           &           & $11$ & $4$ \\
\bottomrule
\end{tabular}
\end{adjustbox}
\end{minipage}
\end{table*}

\paragraph{RQ14: Does DPO refresh degrade the verifier?}
Table~\ref{tab:verifier_drift} compares the base Qwen3-VL-8B verifier with the DPO-updated checkpoint on general benchmarks. All differences are within $\pm0.3$, so DPO refresh preserves general visual understanding. The benefit of the refresh appears in grounding evaluation instead, since freezing the verifier lowers EMMA from $27.3$ to $26.1$ (Table~\ref{tab:ablation_arch}, row 3).

\paragraph{RQ15: Does the policy route queries to the appropriate mode?}
During RL, the adaptive slots select the utility-optimal mode for $40\%$ of queries after SFT and for $75\%$ by the end of training. At evaluation, the utility-optimal mode is unavailable because it requires controlled rollouts, so we use GPT-5.5 as an independent judge of the most appropriate mode for all $1{,}837$ Video-Holmes queries. \modelname agrees with the judge on $72\%$ of queries. Table~\ref{tab:holmes_routing} partitions the queries by agreement and forces the judged mode on the queries where the two disagree. Forcing the judged mode raises accuracy on these queries from $40$ to $47$ and recovers $7$ of the $11$-point gap, so routing is the main source of the gap and the residual $4$ points reflect genuinely harder queries. Mis-routing concentrates in Social Reasoning, where identity queries look perceptual but require tracking across frames (\S\ref{app:limitations}).

\section{Full Training Algorithm}
\label{app:algorithm}

Algorithm~\ref{alg:training} gives a self-contained procedural view of \modelname.

\newcommand{\algcmt}[1]{\hfill{\small\textcolor{gray!65!black}{$\triangleright$\,\textit{#1}}}}

\begin{algorithm}[t]
\caption{\modelname~Full Training Procedure}
\label{alg:training}
\begin{algorithmic}[1]

\REQUIRE Base VLM $\pi_\theta$ (Qwen3-VL); SFT data $\mathcal{D}_{\mathrm{SFT}}$;
         RL corpus $\mathcal{D}_{\mathrm{RL}}$ (medium-difficulty); $K{=}8$, $\delta{=}0.35$,
         $\beta_{\mathrm{alp}}{=}0.20$; verifier $V$; refresh interval $N_{\mathrm{ref}}$, total RL steps $T_{\mathrm{RL}}$
\ENSURE  Adaptive policy $\pi_\theta^*$ that selects reasoning mode per query

\STATE Fine-tune $\pi_\theta$ on mode-labeled traces from $\mathcal{D}_{\mathrm{SFT}}$:

\FOR{step $t = 1$ \textbf{to} $T_{\mathrm{RL}}$}

  \STATE Sample batch $\mathcal{B}$ from $\mathcal{D}_{\mathrm{RL}}$
  \vspace{1pt}

  \STATE \textcolor{gray!65!black}{\textit{\small\textbf{(a)} Mode-structured rollout generation}}
  \FORALL{prompt $q \in \mathcal{B}$}
    \STATE Generate $K{=}8$ completions with fixed slot assignment: \algcmt{controlled $\mid$ adaptive}
    \STATE \hspace{1.3em}
      $\bigl[$\,
      {\color{directcolor}$\mathbf{D}_1\mathbf{D}_2$},\;
      {\color{concisecolor}$\mathbf{C}_3\mathbf{C}_4$},\;
      {\color{deepcolor}$\mathbf{P}_5\mathbf{P}_6$},\;
      {\color{adaptivecolor}$\mathbf{A}_7\mathbf{A}_8$}\,
      $\bigr]$
    \STATE Inject mode instruction per slot; all $K$ scored under identical reward
  \ENDFOR
  \vspace{1pt}

  \STATE \textcolor{gray!65!black}{\textit{\small\textbf{(b)} Per-rollout reward computation}}
  \FORALL{rollout $o_j$ in batch}
    \STATE {\color{darkgreen}$R_{\mathrm{ans}}$}: exact-match / soft-numeric / ROUGE-L
           \hspace{0.4em}{\color{darkgreen}$R_{\mathrm{format}}$}: schema validity
           \hspace{0.4em}{\color{darkgreen}$R_{\mathrm{comply}}$}: compliance + overflow
    \STATE $R_{\mathrm{balance}}$: penalize mode collapse across the $K$-group
           \algcmt{maintains diversity}
    \STATE {\color{deepcolor}$R_{\mathrm{cost}}$} $= -\!\bigl(\mathrm{BaseCost}(m)
           + \beta_{\mathrm{alp}}\,\tfrac{|\hat{y}|}{L_0}\,\hat{p}_{\mathrm{solved}}\,\mu_m\bigr)$
           \algcmt{ALP: harder prompts penalized less}
    \IF{$j \in \{7,8\}$ \textbf{({\color{adaptivecolor}adaptive} slots)}}
      \STATE From controlled rollouts $\{o_1,\ldots,o_6\}$ compute per-mode utility:
      \STATE \hspace{1.3em}
             $\mathrm{Utility}(m) = \!\bigl(\mathrm{Ans}(m)+\delta\,G(m)\bigr)
               \bigl(1-\mathrm{Cost}(m)\bigr)$
      \STATE \hspace{1.3em}
             $m^* = \arg\max_m\,\mathrm{Utility}(m)$
             \algcmt{ties broken toward cheaper mode}
      \STATE {\color{adaptivecolor}$R_{\mathrm{select}}$}:
             $+\Delta$ if $o_j$ chose $m^*$;\quad $-\Delta$ otherwise
    \ENDIF
    \STATE Query verifier $V$ $\Rightarrow$
           {\color{cvprblue}$R_{\mathrm{verifier}}$} $= w_t v_t + w_s v_s + w_h v_h + \cdots$
           \algcmt{temporal, spatial, human-alignment; span/tag/pref/replay terms in \S\ref{app:reward_ver}}
  \ENDFOR
  \vspace{1pt}

  \STATE \textcolor{gray!65!black}{\textit{\small\textbf{(c)} Token-level credit shaping}}
  \STATE Obtain per-token credit $c_t$ from verifier spans (heuristic fallback if absent)
  \STATE Reshape advantages: $\hat{A}_t = A_{\mathrm{seq}}\cdot c_t$, \;
         $\tfrac{1}{T}\sum_t c_t = 1$
         \algcmt{evidence spans receive up to $4\times$ gradient}
  \vspace{1pt}

  \STATE \textcolor{gray!65!black}{\textit{\small\textbf{(d)} GDPO normalization + DAPO update}}
  \STATE $A_k^{(i,j)} \leftarrow \tfrac{r_k^{(i,j)}-\mu_k}{\sigma_k+\epsilon}$ for each $r_k$;\quad
         $A_{\mathrm{sum}} = \textstyle\sum_{k=1}^{7}A_k^{(i,j)}$
  \STATE Batch-renormalize $A_{\mathrm{sum}}$;\quad update $\pi_\theta$ via DAPO objective \eqref{eq:dapo}
  \vspace{1pt}

  \IF{$t \bmod N_{\mathrm{ref}} = 0$}
    \STATE \textcolor{gray!65!black}{\textit{\small\textbf{(e)} Online verifier DPO refresh}}
    \STATE Build preference pairs: chosen $=$ highest-$R$ rollout;\; rejected $=$ lowest-$R$
    \STATE DPO fine-tune $V \xrightarrow{\;\mathcal{P}_{\mathrm{pref}}\;} V'$;\quad
           hot-swap $V' \to R_{\mathrm{verifier}}$
           \algcmt{keeps verifier calibrated}
  \ENDIF

\ENDFOR
\RETURN $\pi_\theta$

\end{algorithmic}
\end{algorithm}

\section{Evaluation Details}
\label{app:eval_details}

\subsection{Benchmarks and Splits}
\label{app:eval_benchmarks}

We evaluate on a diverse suite of multimodal benchmarks spanning perception-heavy multiple-choice tasks, open-ended visual reasoning, video temporal grounding, and knowledge-intensive video QA.

\subsection{Image Benchmarks}

\textbf{EMMA}~\cite{hao2025emma} evaluates organic multimodal reasoning in mathematics, physics, chemistry, and coding, where text-only chain-of-thought is insufficient, and the visual context is load-bearing.
It probes whether a model can genuinely integrate visual and textual information in a back-and-forth manner rather than treating the image as supplementary context.

\textbf{MMMU}~\cite{yue2024mmmu} contains 11,500 college-level questions across 30 subjects and 183 subfields, requiring expert domain knowledge fused with heterogeneous visual inputs (diagrams, tables, charts).
Its breadth and disciplinary depth make it a strong test of models that must balance deep deliberation on hard items with efficiency on simpler perceptual lookups.

\textbf{MathVista}~\cite{lu2024mathvista} covers seven mathematical reasoning types (algebraic, arithmetic, geometric, logical, statistical, and more) that are tightly coupled with diverse visual contexts, such as plots, geometric figures, and tables.
It tests visually grounded arithmetic and multi-step inference, where the answer cannot be derived from language priors alone.

\textbf{HallusionBench}~\cite{guan2024hallusionbench} is specifically designed to disentangle language hallucination (answering from memory) and visual illusion (misreading the image), using matched control-group question pairs to isolate each failure mode.
It is selected to evaluate whether grounding rewards suppress the tendency to hallucinate rather than read the visual signal directly.

\textbf{AI2D}~\cite{kembhavi2016ai2d} presents grade-school science diagrams (food webs, water cycles, planetary systems) with structural annotations and multiple-choice questions.
It covers non-photographic structured graphics where models must parse constituent relationships and apply domain semantics, going beyond recognition of natural images.

\textbf{MM-Vet~v2}~\cite{yu2024mmvetv2} evaluates seven integrated capabilities, including recognition, knowledge, OCR, spatial awareness, language generation, math, and image-text sequence understanding across interleaved multi-image inputs.
Its open-ended, judge-scored format captures response quality beyond accuracy, complementing the multiple-choice benchmarks in our suite.

\subsection{Video Benchmarks}

\textbf{Video-Holmes}~\cite{videoholmes} requires models to actively locate and causally chain multiple visual clues scattered across temporal segments in suspense short films, with even the strongest models (Gemini-2.5-Pro) achieving only $\sim$45\% accuracy.
It tests multi-hop inferential reasoning, in which evidence must be retrieved and linked over time rather than recalled from a single frame.

\textbf{Video-TT}~\cite{zhang2025videott} evaluates both correctness and robustness of video-LLMs using 1,000 YouTube Shorts paired with adversarial question variants (rephrased, wrongly-led, correctly-led).
The robustness track is sensitive to over-reasoning and confirmation bias, testing whether models can maintain accurate answers under natural adversarial pressure.

\textbf{VSI-Bench}~\cite{yang2025vsibench} tests visual-spatial intelligence through 5,000+ QA pairs over indoor-scene videos, covering spatial measurement, distance estimation, object counting, and route planning.
It reveals that spatial reasoning, not perception or language, is the primary bottleneck for current models, making it a useful probe of grounded inference over sequential frames.

\textbf{SciVideoBench}~\cite{deng2025scivideobench} presents 1,000 research-level questions from experimental videos across 25+ specialized domains (physics, chemistry, biology, medicine).
It sits well beyond saturated college-level benchmarks, requiring the combination of precise spatiotemporal perception, expert domain knowledge, and sophisticated logical reasoning.

\textbf{Video-MMMU}~\cite{hu2025videommmu} measures knowledge acquisition from 300 expert educational videos through 900 questions aligned to Bloom's taxonomy (Perception, Comprehension, Adaptation).
Unlike static benchmarks, it requires models to learn and apply knowledge from video content during inference, with a steep drop in performance as cognitive demand increases.

\section{Reward Function and Utility: Full Definitions}
\label{app:reward_math}
This section provides complete mathematical definitions for each reward component introduced in \S\ref{sec:reward} and \S\ref{sec:utility}, including per-answer-type scoring rules, cost terms, and the utility function used to supervise adaptive slot selection.

\subsection{\texorpdfstring{Answer Reward $R_{\mathrm{ans}}$}{Answer Reward R\_ans}}
\label{app:reward_ans}
Answer scoring is dispatched by the \texttt{answer\_type} field attached to each training example.
Let $\hat{a}$ denote the extracted prediction and $a^*$ the gold solution.

\paragraph{mcq\_letter.}
Exact uppercase letter match:
\begin{equation}
  R_{\mathrm{ans}} = \mathbb{1}[\hat{a} = a^*] \in \{0, 1\}.
\end{equation}

\paragraph{number.}
Soft inverse-MSE score that degrades smoothly with numeric error:
\begin{equation}
  R_{\mathrm{ans}} = \frac{1}{1 + (\hat{a} - a^*)^2}.
\end{equation}

\paragraph{temporal\_single.}
A single predicted timestamp $\hat{t}$ is scored against the gold timestamp $t^*$ with a tolerance-normalized soft score:
\begin{equation}
  R_{\mathrm{ans}} = \max\!\bigl(0,\, 1 - \tfrac{|\hat{t} - t^*|}{\tau_0}\bigr), \quad \tau_0 = 5.0\,\text{s}.
\end{equation}
This gives full credit at the gold timestamp and decays linearly to zero at $\tau_0$ seconds away.

\paragraph{temporal\_span.}
A predicted span $[\hat{t}_1, \hat{t}_2]$ is scored against gold $[t^*_1, t^*_2]$ with temporal IoU (tIoU):
\begin{equation}
  R_{\mathrm{ans}} = \mathrm{tIoU}(\hat{a}, a^*) = \frac{\max(0,\,\min(\hat{t}_2, t^*_2) - \max(\hat{t}_1, t^*_1))}{\max(\hat{t}_2, t^*_2) - \min(\hat{t}_1, t^*_1)}.
\end{equation}
Disjoint predictions receive $R_{\mathrm{ans}} = 0$.

\paragraph{bbox.}
A predicted bounding box $\hat{b} = [x_1, y_1, x_2, y_2]$ is scored against gold $b^*$ with standard 2D intersection-over-union:
\begin{equation}
  R_{\mathrm{ans}} = \mathrm{IoU}(\hat{b},\, b^*) = \frac{|\hat{b} \cap b^*|}{|\hat{b} \cup b^*|} \in [0, 1].
\end{equation}

\paragraph{ocr/open/caption.}
Plain-text outputs are scored with ROUGE-L $F_1$:
\begin{equation}
  R_{\mathrm{ans}} = \mathrm{ROUGE\text{-}L}_{F_1}(\hat{a},\, a^*).
\end{equation}

ROUGE-L is the standard metric used by the corresponding benchmarks, ensuring that the answer reward aligns with the evaluation protocol rather than introducing a new design choice.

\subsection{\texorpdfstring{Format Reward $R_{\mathrm{format}}$}{Format Reward R\_fmt}}
A binary reward:
\begin{equation}
  R_{\mathrm{format}} = \mathbb{1}\bigl[\text{completion matches full schema regex}\bigr] \in \{0, 1\}.
\end{equation}
The schema regex requires: leading \texttt{<MODE>} tag, \texttt{<think>} block, \texttt{<answer>} block, matching closing \texttt{</MODE>} tag, and no content outside the wrapper after stripping trailing chat special tokens.

\subsection{\texorpdfstring{Compliance Reward $R_{\mathrm{comply}}$}{Compliance Reward R\_comp}}
For a completion $c$ assigned mode $m$:
\begin{equation}
  R_{\mathrm{comply}} = \begin{cases}
    -2.0 & \text{if format invalid (schema regex fails)},\\
    -2.0 & \text{if assigned mode} \neq \text{produced mode (controlled slots)},\\
    r_{\text{ok}} - \lambda_L \cdot \max\!\bigl(0,\,\tfrac{w_c - w_m}{w_m}\bigr) & \text{otherwise},
  \end{cases}
\end{equation}
where $r_{\text{ok}} = 0.5$, $\lambda_L = 0.5$, $w_c$ is the \texttt{<think>} word count, and $w_m$ is the per-mode hard limit ($w_{\texttt{DIRECT}} = 120$, $w_{\texttt{CONCISE}} = 600$, no explicit cap for \tagDeep ).
Adaptive-slot completions (slot index $\geq 7$) skip the mode-mismatch check but still receive the format-validity and length-overflow checks.

\subsection{\texorpdfstring{Balance Reward $R_{\mathrm{balance}}$}{Balance Reward R\_bal}}
While $R_{\mathrm{comply}}$ fires \emph{per-completion} when a single rollout ignores its assigned mode, $R_{\mathrm{balance}}$ acts at the \emph{group level}: it corrects under-production of any one mode across the six controlled slots, a failure that $R_{\mathrm{comply}}$ alone cannot prevent (a rollout that produces the right mode tag but in the wrong proportion contributes nothing to $R_{\mathrm{comply}}$ yet degrades the within-group utility estimate).
For a group $\mathcal{G}$ of $K$ rollouts, let $n_m$ be the count of controlled-slot completions with produced mode $m$ and $t_m$ be the target count ($t_{\texttt{DIRECT}} = t_{\texttt{CONCISE}} = t_{\texttt{DEEP}} = 2$ for $K=8$).
For a controlled-slot rollout at index $i$ with produced mode $m_i$:
\begin{equation}
  R_{\mathrm{balance},i} = \mathrm{clip}\!\left(b_{m_i} + \frac{t_{m_i} - n_{m_i}}{t_{m_i}},\, -1,\, 1\right), \quad
  b_m = \begin{cases} 0.2 & m \in \{D,C,P\},\\ -0.2 & \text{otherwise.} \end{cases}
\end{equation}
Adaptive-slot rollouts (slot index $\geq 7$) receive $R_{\mathrm{balance}} = 0$.

\subsection{\texorpdfstring{Cost and Length Reward $R_{\mathrm{cost}}$}{Cost and Length Reward R\_cost}}
Let $\mathrm{cost}_{\mathrm{eff}}(m, c)$ denote the effective mode cost:
\begin{equation}
  \mathrm{cost}_{\mathrm{eff}}(m,c) = C_m + \lambda_O \cdot \max\!\bigl(0,\,\tfrac{w_c - W_m}{W_m}\bigr),
\end{equation}
where $w_c$ is the word count of completion $c$, $C_m$ is the base mode cost ($0.00 / 0.08 / 0.20$ for \tagDirect /\tagConcise /\tagDeep), $\lambda_O = 0.25$ is the overflow scale, and $W_m$ is the per-mode word budget ($120$ and $600$ words for \tagDirect and \tagConcise, with no hard cap for \tagDeep).
The base penalty is:
\begin{equation}
  P_{\mathrm{base}} = \mathrm{cost}_{\mathrm{eff}}(m,c) \cdot \bigl[\alpha_w + (1-\alpha_w)\,R_{\mathrm{ans}}\bigr], \quad \alpha_w = 0.25.
\end{equation}
The ALP penalty (see Eq.~\eqref{eq:alp} in the main text) is:
\begin{equation}
  P_{\mathrm{alp}} = \beta_{\mathrm{alp}}\,\frac{|\hat{y}|}{L_0}\,\hat{p}_{\mathrm{solved}}\,\mu_m,
\end{equation}
where $\beta_{\mathrm{alp}} = 0.20$, $L_0 = 512$ is a reference length for normalization, $\hat{p}_{\mathrm{solved}}$ is the fraction of group rollouts with $R_{\mathrm{ans}} \geq 0.999$ clipped to a minimum of $1/K$, and $\mu_m \in \{1.00,1.10,1.25,1.10\}$ for (\tagDirect, \tagConcise, \tagDeep, \tagAdaptive).
The full cost reward is $R_{\mathrm{cost}} = -(P_{\mathrm{base}} + P_{\mathrm{alp}})$.

\subsubsection{Cost Pressure: Scenario Analysis}
\label{app:cost_analysis}

$R_{\mathrm{cost}} = -(P_{\mathrm{base}} + P_{\mathrm{alp}})$ must be \emph{adaptive} (scale with query difficulty via $\hat{p}_{\mathrm{solved}}$) and \emph{stable} (work before $\hat{p}_{\mathrm{solved}}$ is reliable).
Table~\ref{tab:utility_cases} shows which mode wins under $\mathrm{Utility}(m)=\bigl(\mathrm{Ans}(m)+\delta\cdot G(m)\bigr)\bigl(1-\mathrm{Cost}(m)\bigr)$, where $\mathrm{Ans}(m)$ is the mean $R_{\mathrm{ans}}$ over controlled rollouts for mode $m$, $G(m)$ is the mean verifier grounding score in $[0,1]$, and $\mathrm{Cost}(m)$ is the base mode cost $C_m$; Table~\ref{tab:cost_design} summarizes when each design component is needed.
Note: $\hat{p}_{\mathrm{solved}}$ is computed over all $K=8$ slots, so it tracks query difficulty rather than per-mode success.

\begin{table*}[!t]
\caption{Representative utility outcomes ($\delta=0.35$). $\mathrm{Ans}(m)$ is the mean $R_{\mathrm{ans}}$ over the two controlled rollouts for mode $m$; $G(m) = \tfrac{1}{2}(v_t(m)+v_s(m))$ is the mean verifier grounding score ($v_t, v_s \in [0,1]$, see \S\ref{app:verifier_interface}); $\mathrm{Cost}(m)$ is the base mode cost $C_m \in \{0.00, 0.08, 0.20\}$ for \textsc{direct}/\textsc{concise}/\textsc{deep} (overflow term omitted for illustration). Bold = $m^*$.}
\label{tab:utility_cases}
\smallskip
\begin{adjustbox}{max width=\textwidth,center}
\renewcommand{\arraystretch}{1.2}
\begin{tabular}{l l c c c c c}
\toprule
Difficulty & Mode & $\mathrm{Ans}(m)$ & $G(m)$ & $\mathrm{Cost}(m)$ & Utility & $m^*$ \\
\midrule
\multirow{3}{*}{Easy}
  & \tagDirect   & 0.95 & 0.30 & 0.00 & $\mathbf{1.055}$ & \checkmark \\
  & \tagConcise & 0.90 & 0.50 & 0.08 & 0.989 & \\
  & \tagDeep         & 0.88 & 0.55 & 0.20 & 0.858 & \\
\midrule
\multirow{3}{*}{Medium}
  & \tagDirect    & 0.55 & 0.20 & 0.00 & 0.620 & \\
  & \tagConcise & 0.82 & 0.65 & 0.08 & $\mathbf{0.964}$ & \checkmark \\
  & \tagDeep           & 0.80 & 0.70 & 0.20 & 0.836 & \\
\midrule
\multirow{3}{*}{Hard}
  & \tagDirect  & 0.25 & 0.10 & 0.00 & 0.285 & \\
  & \tagConcise & 0.42 & 0.50 & 0.08 & 0.547 & \\
  & \tagDeep          & 0.72 & 0.78 & 0.20 & $\mathbf{0.794}$ & \checkmark \\
\bottomrule
\end{tabular}
\end{adjustbox}
\end{table*}

\paragraph{Easy query (\tagDeep chosen incorrectly), $\hat{p}_{\mathrm{solved}}=0.75$.}
On an easy retrieval query, \tagDirect and \tagConcise controlled slots both succeed reliably, keeping $\hat{p}_{\mathrm{solved}}$ high regardless of \tagDeep's outcome.
A \tagDeep rollout with $|\hat{y}|=600$ tokens receives:
\begin{align}
  P_{\mathrm{alp}} &= 0.20\cdot\tfrac{600}{512}\cdot 0.75\cdot 1.25 = 0.220, \\
  P_{\mathrm{base}} &= 0.20\cdot[0.25+0.75\cdot0.95] = 0.193, \\
  R_{\mathrm{cost}} &= -(0.220+0.193) = -0.413.
\end{align}
The same query under \tagDirect with $|\hat{y}|=80$ tokens:
\begin{align}
  P_{\mathrm{alp}} &= 0.20\cdot\tfrac{80}{512}\cdot 0.75\cdot 1.00 = 0.023, \\
  P_{\mathrm{base}} &= 0.000 \quad (C_{\mathrm{Direct}}=0), \\
  R_{\mathrm{cost}} &= -0.023.
\end{align}
\tagDirect's cost advantage is $0.413 - 0.023 = 0.390$, creating a strong within-group signal that propagates via GDPO to the adaptive slots.

\paragraph{Hard query (\tagDeep justified), $\hat{p}_{\mathrm{solved}}=0.125$.}
On a hard multi-hop query, most rollouts fail, so $\hat{p}_{\mathrm{solved}}$ is low, and ALP relaxes.
A \tagDeep rollout with $|\hat{y}|=700$ tokens now receives:
\begin{align}
  P_{\mathrm{alp}} &= 0.20\cdot\tfrac{700}{512}\cdot 0.125\cdot 1.25 = 0.043, \\
  P_{\mathrm{base}} &= 0.20\cdot[0.25+0.75\cdot0.72] = 0.158, \\
  R_{\mathrm{cost}} &= -(0.043+0.158) = -0.201.
\end{align}
\tagDirect on the same prompt ($R_{\mathrm{ans}}=0.25$) gets $R_{\mathrm{cost}}=-0.004$, so \tagDirect's cost advantage shrinks to $0.201 - 0.004 = 0.197$.
This is more than offset by \tagDeep's $R_{\mathrm{ans}}$ advantage of $0.72 - 0.25 = 0.47$, so the composite reward still favors \tagDeep.

\begin{table*}[!t]
\caption{Design rationale: which component handles each regime. \checkmark\,=\,correct; \texttimes\,=\,failure.}
\label{tab:cost_design}
\smallskip
\begin{adjustbox}{max width=\textwidth,center}
\renewcommand{\arraystretch}{1.2}
\begin{tabular}{l c c c l}
\toprule
Design & Early training & Easy query & Hard query & Failure mode \\
\midrule
ALP only (no $C_m$)     & \texttimes & \checkmark & \checkmark & $\hat{p}_{\mathrm{solved}}\!\approx\!0$ early $\Rightarrow$ no signal; \tagDeep -as-default collapse \\
\rowcolor{gray!8}
Base cost only (no ALP) & \checkmark & partial    & \texttimes & Difficulty-blind; suppresses \tagDeep  equally on hard tasks \\
\rowcolor{GroupColor1}
Base $+$ ALP (ours)     & \checkmark & \checkmark & \checkmark & --- \\
\bottomrule
\end{tabular}
\end{adjustbox}
\end{table*}

\FloatBarrier

\noindent The specific constants ($C_{\mathrm{Deep}}=0.20$, $\beta_{\mathrm{alp}}=0.20$, $\mu_{\mathrm{Deep}}=1.25$) are chosen so that on a typical easy query ($\hat{p}_{\mathrm{solved}}\approx0.8$, $|\hat{y}|\approx600$) the total \tagDeep penalty is $0.35$--$0.45$, exceeding the \tagDirect--\tagDeep accuracy gap on retrieval benchmarks and falling below it on disambiguation benchmarks, producing the bifurcation in $m^*$ visible in Table~\ref{tab:utility_cases}.

\subsection{\texorpdfstring{Selection Reward $R_{\mathrm{select}}$}{Selection Reward R\_sel}}
For each group $\mathcal{G}$, compute mode utilities from controlled slots:
\begin{equation}
  \mathrm{Utility}(m) = \bigl(\overline{R_{\mathrm{ans}}}(m) + \delta\,\overline{G}(m)\bigr)\bigl(1 - \overline{\mathrm{cost}_{\mathrm{eff}}}(m)\bigr),
\end{equation}
where $\overline{R_{\mathrm{ans}}}(m)$, $\overline{G}(m) = \tfrac{1}{2}(\overline{v_t}(m) + \overline{v_s}(m))$, and $\overline{\mathrm{cost}_{\mathrm{eff}}}(m)$ are means over the 2 controlled rollouts for mode $m$, $\delta = 0.35$. Here $v_t, v_s \in [0,1]$ are the temporal and spatial grounding scores returned by the verifier (full definitions in Appendix~\S\ref{app:verifier_interface}).
$\delta$ is chosen so that grounding breaks near-ties in mode selection without allowing a grounded incorrect response to outscore an accurate ungrounded one: too small and grounding has no influence on $m^*$, too large and correctness is no longer the dominant signal.
Sensitivity to $\delta \in \{0.35, 0.70\}$ is reported in Table~\ref{tab:sensitivity}.
$m^* = \arg\max_{m} \mathrm{Utility}(m)$ (ties broken toward the cheaper mode).
For adaptive rollout $i$ with chosen mode $m_i$:
\begin{equation}
  R_{\mathrm{select},i} = \begin{cases}
    +0.8 & \text{if } m_i = m^*,\\
    -0.8 & \text{if } m_i \neq m^* \text{ or } m_i \notin \{\texttt{D,C,P}\},\\
    0.0  & \text{if group has no valid controlled utility estimate.}
  \end{cases}
\end{equation}
The selection reward requires the full group to be available. For distributed training, an all-gather over ranks collects all completions before scoring.

\subsection{\texorpdfstring{Verifier Reward $R_{\mathrm{verifier}}$}{Verifier Reward R\_ver}}
\label{app:reward_ver}
\label{app:verifier_interface}
\label{app:verifier_dpo}

The verifier (Qwen3-VL-8B) scores each rollout on three dimensions ($v_t$, $v_s$, $v_h$) and returns span-level annotations used by $R_{\mathrm{verifier}}$ and token-level credit shaping (\S\ref{app:token_credit}); the preferred-mode signal is derived from these scores:

\begin{itemize}
  \item \textbf{Temporal grounding} ($v_t \in [0,1]$): whether \texttt{<think>} contains explicit \texttt{<t>} time anchors with consistent ordering; vague temporal references without timestamps are penalized.
  \item \textbf{Spatial grounding} ($v_s \in [0,1]$): whether \texttt{<obj>}+\texttt{<box>} pairs appear with plausible correspondence; object mentions without localization are penalized.
  \item \textbf{Human alignment} ($v_h \in [0,1]$): whether reasoning is internally consistent, evidence-linked, and free of repetitive filler or unjustified certainty. Enforcing alignment increases response length and thus token cost; $R_{\mathrm{cost}}$ directly accounts for this overhead, making the signal interpretable without penalizing depth when warranted.
  \item \textbf{Preferred mode}: the lowest-cost mode judged reliable for this sample under the ordering \tagDirect $<$ \tagConcise $<$ \tagDeep. Adaptive slots whose chosen mode conflicts with this judgment receive a corresponding penalty.
  \item \textbf{Span scores} ($S_{\mathrm{span}} \in [-1,1]$ per sentence): positive for evidence-supporting spans, negative for filler or hallucinated content, used directly as token-level credit weights when available, otherwise computed from the local heuristic (\S\ref{app:token_credit}).
\end{itemize}

\paragraph{Reward aggregation.}
The scalar verifier reward aggregates the signals above:
\begin{equation}
  R_{\mathrm{verifier}} = w_t v_t + w_s v_s + w_h v_h + w_{\mathrm{span}} S_{\mathrm{span}} + \Delta_{\mathrm{tag}} + \Delta_{\mathrm{pref}} + \Delta_{\mathrm{replay}},
\end{equation}
where:
\begin{itemize}
  \item $(w_t, w_s, w_h) = (0.35, 0.35, 0.30)$,
  \item $w_{\mathrm{span}} = 0.25$,
  \item $\Delta_{\mathrm{tag}} = -0.6$ if mode is \tagDirect and any grounding tag (\texttt{<obj>}, \texttt{<box>}, \texttt{<t>}) appears in \texttt{<think>}, else $0$,
  \item $\Delta_{\mathrm{pref}} = +0.4$ if adaptive slot chose verifier's \texttt{preferred\_mode}, else $-0.4$ for adaptive slots, $0$ for controlled slots,
  \item $\Delta_{\mathrm{replay}} = 0.15 \cdot \min(p_{\mathrm{old}}/2, 1)$ where $p_{\mathrm{old}} \in [0,3]$ is the replay priority for this prompt (see Appendix~\S\ref{app:replay}).
\end{itemize}
The full reward is clipped to $[-2.0, 2.0]$.

\paragraph{Online DPO refresh.}
Every $N=100$ RL gradient steps, the highest- and lowest-reward completions within each rollout group form a DPO preference pair (chosen/rejected), which is retained if the reward margin exceeds $0.15$.
Up to 2{,}000 pairs are constructed per trigger and used to fine-tune the verifier via DPO, after which the updated checkpoint is hot-swapped into the reward pipeline.
This keeps the verifier calibrated to the current policy without interrupting RL training.

\paragraph{Verifier identity and drift.} The verifier is initialized from the base Qwen3-VL-8B checkpoint, not the SFT policy, and receives no gradients from the policy loss. DPO refresh uses rollouts from the current policy, but pairs are ranked by the composite reward in which $R_{\mathrm{ans}}$ dominates, so a rollout with high verifier scores and a wrong answer is rejected. DPO therefore learns to distinguish grounding quality among correct rollouts. The refresh also preserves the verifier's general capability (Table~\ref{tab:verifier_drift}, RQ14).

\section{Token-Level Credit Construction}
\label{app:token_credit}

\paragraph{Motivation.}
Standard GRPO/DAPO assigns the same advantage to every token in a completion.
For reasoning traces, this dilutes the gradient signal across low-information spans (hedging phrases, transitional filler, repeated prefixes) that are unlikely to distinguish good from bad completions.
Hence, the credit map concentrates the gradient on evidence-bearing spans.

\paragraph{Local heuristic.}
In the absence of verifier \texttt{span\_scores}, the reward function computes a local span-credit score $S_{\mathrm{span}}$ per completion:
\begin{enumerate}
  \item Extract the \texttt{<think>} block, split into sentence-like spans at newline, period, exclamation, or question-mark boundaries.
  \item For each span $s_j$, accumulate:
    \begin{itemize}
      \item $+0.20$ if $s_j$ contains both \texttt{<obj>} and \texttt{<box>} (spatially grounded span).
      \item $+0.15$ if $s_j$ contains \texttt{<t>} (temporally grounded span).
      \item $+0.10$ if $s_j$ has $\geq 4$ non-tag words (substantive text span).
      \item $-0.12$ if $s_j$ contains hedging phrases (``maybe'', ``probably'', ``guess'', ``unsure'', ``not sure'').
    \end{itemize}
  \item Average across spans, clip to $[-1.0, 1.0]$.
\end{enumerate}
Empty think blocks receive $S_{\mathrm{span}} = -0.2$ as a baseline penalty.

\paragraph{Verifier span scores.}
When \texttt{need\_span\_credit=1} is included in the verifier request, the server returns \texttt{span\_scores}: a list of floats in $[-1,1]$ aligned to the span boundaries provided in \texttt{span\_texts}.
These scores replace the local heuristic and provide a model-based assessment of each span's evidential value.

\paragraph{Token-level projection.}
The credit map $c_t$ for each completion token $t$ is constructed by interpolating the span score of the span to which $t$ belongs into a normalized weight:
\begin{equation}
  c_t = \frac{1 + s_{\sigma(t)}}{\sum_{t'=1}^{T} (1 + s_{\sigma(t')}) / T},
\end{equation}
where $\sigma(t)$ is the span index of token $t$ and $s_{\sigma(t)} \in [-1,1]$ is the span score.
The normalization ensures $\tfrac{1}{T}\sum_t c_t = 1$, preserving the overall gradient magnitude.
Tokens outside the \texttt{<think>} block (the \texttt{<answer>} block and wrapper tokens) receive neutral credit ($c_t = 1$) unless explicitly overridden.

\clearpage
\section{Hyperparameters}
\label{app:hyperparams}
\phantomsection\label{app:training_details}
Table~\ref{tab:hyperparams} lists the hyperparameters of \modelname by training component, with the expected effect of raising or lowering each value where relevant.
\begin{table*}[!h]
\centering
\caption{\textbf{Complete hyperparameter listing for \modelname.} Sensitivity sweeps for all parameters were not feasible within our compute budget; $\delta$ and $w_{\mathrm{select}}$ are swept in Table~\ref{tab:sensitivity} and remaining entries document the expected directional effect.}
\label{tab:hyperparams}
\tiny
\setlength{\tabcolsep}{2pt}
\renewcommand{\arraystretch}{0.95}
\begin{adjustbox}{max width=\linewidth,center}
\begin{tabular}{p{1.4cm}p{3.2cm}p{2.2cm}p{5.5cm}}
\toprule
\textbf{Group} & \textbf{Parameter} & \textbf{Value} & \textbf{Sensitivity} \\
\midrule
\multirow{6}{*}{Training}
  & Backbone          & Qwen3-VL / Qwen2.5-VL & --- \\
  & LoRA target       & all-linear             & --- \\
  & Precision         & bfloat16               & --- \\
  & Learning rate     & $1\times10^{-6}$       & --- \\
  & Warmup ratio      & 0.05                   & --- \\
  & Grad.\ accum.\ steps & 2                  & --- \\
\midrule
\multirow{5}{*}{Rollout}
  & $K$ (rollouts per prompt) & 8           & $\uparrow$ better utility estimates, less adaptive practice; $\downarrow$ noisier $\hat{p}_{\mathrm{solved}}$ \\
  & Temperature       & 1.0                    & --- \\
  & Top-$p$           & 1.0                    & --- \\
  & Max context       & 16{,}384 tokens        & --- \\
  & Max completion    & 1{,}024 tokens         & --- \\
\midrule
\multirow{2}{*}{DAPO}
  & $\varepsilon_{\mathrm{low}}$ (lower clip) & 0.20 & DAPO defaults \\
  & $\varepsilon_{\mathrm{high}}$ (upper clip) & 0.28 & DAPO defaults \\
\midrule
\multirow{7}{*}{Reward weights}
  & $w_{\mathrm{ans}}$      & 1.00 & Anchor; scaling equivalent to rescaling all others \\
  & $w_{\mathrm{format}}$   & 0.25 & $\uparrow$ format overpowers task learning; $\downarrow$ schema violations persist \\
  & $w_{\mathrm{comply}}$   & 0.60 & $\uparrow$ mode enforcement dominates; $\downarrow$ forced slots ignore assigned mode \\
  & $w_{\mathrm{balance}}$  & 0.20 & $\uparrow$ forces equal mode mix on easy queries; $\downarrow$ mode collapse risk \\
  & $w_{\mathrm{cost}}$     & 0.20 & $\uparrow$ \tagDeep suppressed on hard queries; $\downarrow$ no efficiency gain \\
  & $w_{\mathrm{select}}$   & 0.40 & Table~\ref{tab:sensitivity} \\
  & $w_{\mathrm{verifier}}$ & 0.35 & $\uparrow$ verifier overrides utility $m^*$; $\downarrow$ grounding feedback negligible \\
\midrule
\multirow{6}{*}{Mode budgets}
  & \tagDirect\ target / hard limit  & 60 / 120 words      & --- \\
  & \tagConcise\ target / hard limit & 200 / 600 words     & --- \\
  & \tagDeep\ target / hard limit    & 1{,}000 words / none & --- \\
  & Base mode cost $C_m$ (D/C/P)     & 0.00 / 0.08 / 0.20 & $\uparrow$ $C_{\mathrm{Deep}}$: \tagDeep suppressed on hard queries; $\downarrow$ over-invoked on easy \\
  & Overflow scale $\lambda_O$       & 0.25                & $\uparrow$ traces aggressively truncated; $\downarrow$ budget overruns tolerated \\
  & Compliance base reward $r_{\mathrm{ok}}$ & 0.5       & --- \\
\midrule
\multirow{5}{*}{ALP / Cost}
  & $\beta_{\mathrm{alp}}$            & 0.20               & $\uparrow$ \tagDeep over-penalized on hard queries; $\downarrow$ too weak to separate modes on easy \\
  & $L_0$ (reference length)          & 512 tokens         & Normalization constant; monotone with $\beta_{\mathrm{alp}}$ \\
  & $\mu_m$ (D/C/P/Adaptive)          & 1.00/1.10/1.25/1.10 & $\uparrow$ ratio: steeper mode cost differentiation; scaling absorbed by $\beta_{\mathrm{alp}}$ \\
  & $\alpha_w$ (base cost weight)     & 0.25               & $\uparrow$ cost more constant (less dependent on correctness); $\downarrow$ cost suppressed on wrong rollouts \\
  & $\lambda_L$ (length penalty)      & 0.50               & --- \\
\midrule
\multirow{2}{*}{Utility}
  & $\delta$ (grounding weight)      & 0.35  & Table~\ref{tab:sensitivity} \\
  & $R_{\mathrm{select}}$ bonus/penalty & $\pm$0.8 & $\uparrow$ selection signal drowns verifier; $\downarrow$ mode routing learns slowly \\
\midrule
\multirow{5}{*}{Verifier}
  & $(w_t, w_s, w_h)$                 & (0.35, 0.35, 0.30) & Shifting ratio redistributes grounding emphasis; near-equal weights chosen \\
  & $w_{\mathrm{span}}$               & 0.25               & $\uparrow$ span credit dominates verifier score; $\downarrow$ token shaping negligible \\
  & $\Delta_{\mathrm{tag}}$           & $-$0.6 & $\uparrow$ magnitude: stronger suppression of grounding in \tagDirect \\
  & $\Delta_{\mathrm{pref}}$          & $\pm$0.4 & $\uparrow$ verifier overrides utility for routing; $\downarrow$ preferred-mode signal has no effect \\
  & Enable step                       & 100                & --- \\
\midrule
\multirow{3}{*}{DPO refresh}
  & Min reward margin                 & 0.15               & $\uparrow$ fewer but cleaner pairs; $\downarrow$ noisy preference labels \\
  & Max pairs per trigger             & 2{,}000            & --- \\
  & Refresh interval $N$              & 100 steps          & $\uparrow$ verifier drifts before recalibration; $\downarrow$ excessive overhead \\
\midrule
\multirow{3}{*}{Replay}
  & Buffer capacity                   & 4{,}096            & --- \\
  & Replay bonus $\gamma_r$           & 0.15               & $\uparrow$ hard examples over-represented; $\downarrow$ no difficulty curriculum \\
  & IS clip $(w_{\min}, w_{\max})$    & (0.5, 2.0)         & --- \\
\bottomrule
\end{tabular}
\end{adjustbox}
\end{table*}

\section{SFT Data}
\label{app:data_construction}
The SFT corpus provides mode-structured warm-start data so that the policy enters RL with well-formed \tagDirect /\tagConcise /\tagDeep behaviors.
We process heterogeneous multimodal QA/reasoning sources into a single unified schema and convert each example into a mode-appropriate completion with strict wrappers and normalized answer blocks.
Mode-structured traces are built upon OneThinker~\citep{feng2025onethinker} and Open-O3-Video~\citep{meng2025open}.

\paragraph{Source datasets and mode assignments.}
Table~\ref{tab:sft_sources} summarizes the three mode partitions with approximate sample counts. The partition sizes reflect data availability. \tagConcise is the largest because it covers the broadest range of tasks, including temporal grounding, charts, math, and general video QA, while \tagDeep requires causal, intent, or long-horizon reasoning for which high-quality annotated data is scarce. The SFT ratios do not determine mode selection. During RL every query receives exactly two controlled rollouts per mode (Eq.~\ref{eq:slot_order}), $R_{\mathrm{balance}}$ penalizes mode collapse within groups, and mode selection is supervised by the within-group utility comparison (Eq.~\ref{eq:utility}). SFT therefore teaches how to execute each mode, and RL teaches when to select it. Despite only $\approx$12K \tagDeep samples, removing \tagDeep degrades the hypothesis-testing benchmarks (Table~\ref{tab:reward_ablation}, R4).

\begin{table*}[!t]
\centering
\caption{SFT data partitions by mode.}
\label{tab:sft_sources}
\begin{adjustbox}{max width=\linewidth,center}
\renewcommand{\arraystretch}{1.15}
\small
\setlength{\tabcolsep}{4pt}
\begin{tabular}{lp{8cm}r}
\toprule
\textbf{Mode} & \textbf{Key sources} & \textbf{Count} \\
\midrule
\tagDirect & Elysium~\cite{wang2024elysium}, TrackingNet~\cite{muller2018trackingnet}, Got10k~\cite{huang2019got}, RefSAV~\cite{yuan2025sa2va}, ReVOS~\cite{jin2025interrvos}, RefYTVOS~\cite{bellver2023closer}, MeViS~\cite{ding2023mevis}, COCO2014~\cite{lin2014microsoft}, LLaVA-ST~\cite{li2025llava}, TextVQA~\cite{singh2019towards} & $\approx$39K \\
\tagConcise & MathQA~\cite{amini2019mathqa}, ChartQA~\cite{masry2022chartqa}, Open-Chart, Charades~\cite{sigurdsson2016hollywood}, DiDeMo~\cite{anne2017localizing}, LongVila~\cite{chen2024longvila}, ShareGPT4V~\cite{chen2024sharegpt4v}, VideoEspresso~\cite{han2025videoespresso}, LLaVA-Video~\cite{zhang2024llava}, Open-O3-Video~\cite{meng2025open} & $\approx$80K \\
\tagDeep & CLEVRER~\cite{yi2019clevrer}, Visulogic~\cite{xu2025visulogic}, STAR~\cite{wu2021star}, PerceptionTest~\cite{patraucean2023perceptiontest}, MovieChat~\cite{song2024moviechat}, EgoProcel~\cite{bansal2022my}, HD-EPIC~\cite{perrett2025hd} & $\approx$12K \\
\bottomrule
\end{tabular}
\end{adjustbox}
\end{table*}

\begin{itemize}
  \item \textbf{DIRECT ($\approx$39K samples).}
  Datasets requiring near-zero reasoning: object tracking (\textit{Elysium}~\cite{wang2024elysium}, \textit{TrackingNet}~\cite{muller2018trackingnet}, \textit{Got10k}~\cite{huang2019got}), video object segmentation (\textit{RefSAV}~\cite{yuan2025sa2va}, \textit{ReVOS}~\cite{jin2025interrvos}, \textit{RefYTVOS}~\cite{bellver2023closer}, \textit{MeViS}~\cite{ding2023mevis}), image segmentation (\textit{COCO2014}~\cite{lin2014microsoft}, capped to $\leq$6K to prevent over-weighting static detection over temporal tracking), spatial grounding (\textit{LLaVA-ST}~\cite{li2025llava}, \textit{Train2014}~\cite{lin2014microsoft}), OCR (\textit{TextVQA}~\cite{singh2019towards}).
  \tagDirect traces target $\leq$60 \texttt{<think>} words and emphasize precise localization.

  \item \textbf{CONCISE ($\approx$80K samples).}
  Procedural reasoning tasks: mathematical QA (\textit{MathQA}~\cite{amini2019mathqa}), chart understanding (\textit{ChartQA}~\cite{masry2022chartqa}, \textit{Open-Chart}), temporal moment retrieval (\textit{Charades}~\cite{sigurdsson2016hollywood}, \textit{DiDeMo}~\cite{anne2017localizing}, \textit{LongVila}~\cite{chen2024longvila}), general multimodal description and QA (\textit{ShareGPT4V}~\cite{chen2024sharegpt4v}, \textit{VideoEspresso}~\cite{han2025videoespresso}), general video QA (\textit{LLaVA-Video}~\cite{zhang2024llava}, downsampled to reduce ``chatterbox'' filler behavior).
  \tagConcise traces target $\leq$200 \texttt{<think>} words with linear factual chains.

  \item \textbf{DEEP ($\approx$12.2K samples).}
  High-value ``trap'' datasets requiring causal, logical, or intent reasoning:
  causal prediction (\textit{CLEVRER}~\cite{yi2019clevrer}), visual logic (\textit{Visulogic}~\cite{xu2025visulogic}), action intent (\textit{STAR}~\cite{wu2021star}), perceptual disambiguation (\textit{PerceptionTest}~\cite{patraucean2023perceptiontest}), long-form movie QA (\textit{MovieChat}~\cite{song2024moviechat}), egocentric procedure understanding (\textit{EgoProcel}~\cite{bansal2022my}, \textit{HD-EPIC}~\cite{perrett2025hd}).
  \tagDeep traces target $\leq$1000 \texttt{<think>} words and are rewritten into negative-reasoning style where appropriate: the model is expected to pause (\emph{``Wait\ldots''}) and falsify plausible distractors before committing to an answer.
\end{itemize}

\section{RL Training Setup}
\label{app:rl_setup}

\paragraph{RL difficulty filtering.}
For the RL training split, we apply a parallel difficulty screening procedure.
Each question in the SFT candidate pool is posed to the SFT-initialized policy across $4$ samples at temperature $1.0$.
We discard a question as too easy when the policy answers it correctly in at least $n_{\mathrm{easy}} = 3$ of $4$ samples. Its rollouts then receive nearly identical rewards, so their group-relative advantages are close to zero and the question barely updates the policy.
We also discard a question as too hard when the policy answers it correctly in at most $n_{\mathrm{hard}} = 1$ of $4$ samples, because almost none of its rollouts receive a positive advantage.
The remaining ``medium-difficulty'' questions constitute the RL dataset ($\approx$ 9K samples), preserving examples where the policy sometimes succeeds and sometimes fails, the regime where mode-selection learning has the greatest impact.

\paragraph{Backbone and LoRA configuration.}
We fine-tune Qwen3-VL~\cite{qwen3vl} and Qwen2.5-VL~\cite{qwen2_5_vl} base models with LoRA~\cite{hu2022lora} applied to all linear layers, bfloat16 precision, and DeepSpeed ZeRO-3~\cite{rajbhandari2020zero} for distributed sharding.
The base model path is the SFT-initialized checkpoint produced from the data construction pipeline (\S\ref{app:data_construction}).

\paragraph{Rollout configuration.}
We use $K=8$ rollouts per prompt with temperature $1.0$ and top-$p$ $1.0$.
Maximum context length is 16{,}384 tokens, maximum completion length is 1{,}024 tokens.
Rollout generation uses vLLM~\cite{vllm} in server mode (external vLLM process, communicating over HTTP).

\paragraph{Slot assignment and mode injection.}
The slot order is configured via:
\begin{equation}
  [\underbrace{\tagDirect,\tagDirect}_{\text{slots 1--2}},\,
   \underbrace{\tagConcise,\tagConcise}_{\text{slots 3--4}},\,
   \underbrace{\tagDeep,\tagDeep}_{\text{slots 5--6}},\,
   \underbrace{\tagAdaptive, \tagAdaptive}_{\text{slots 7--8}}].
\end{equation}

Slot indices are assigned by computing the global repeated-sampling position modulo $K$ (not the local shard index), ensuring that all 8 rollouts for a given prompt receive contiguous slots 1--8 regardless of how training examples are sharded across GPUs.
The mode instruction corresponding to each slot is injected into the prompt at sampler time.

\paragraph{Optimization hyperparameters.}
We train for 1 epoch on the RL dataset with a learning rate of $1\times10^{-6}$, a warmup ratio of 0.05, gradient accumulation over 2 steps, and a per-device batch size of 1.
We use the DAPO~\cite{yu2025dapo} loss with GDPO~\cite{liu2026gdpo} reward scaling, which normalizes each reward function's contribution within the group before combining.

\begin{table*}[!t]
\centering
\caption{Training cost of \modelname relative to standard GRPO on 8 L40S and 4 A100 GPUs. Both use $K{=}8$ rollouts per query.}
\label{tab:compute}
\begin{adjustbox}{max width=\linewidth,center}
\renewcommand{\arraystretch}{1.15}
\small
\setlength{\tabcolsep}{4pt}
\begin{tabular}{l|cc|c}
\toprule
\textbf{Component} & \textbf{Standard GRPO} & \textbf{\modelname} & \textbf{Overhead} \\
\midrule
Rollout generation ($K{=}8$, vLLM)              & \checkmark & \checkmark & $1\times$ \\
Policy and reference forward and backward      & \checkmark & \checkmark & $1\times$ \\
Verifier inference ($8$ calls per batch, vLLM) & \texttimes & \checkmark & $+0.6$ hrs \\
DPO refresh (every $100$ steps, $2$K pairs)    & \texttimes & \checkmark & $+1.9$ hrs \\
\midrule
\rowcolor{GroupColor1}
Total wall-clock time                          & $84$ hrs   & $87$ hrs   & $1.04\times$ \\
\bottomrule
\end{tabular}
\end{adjustbox}
\end{table*}

\paragraph{Training cost.} \modelname and standard GRPO both generate $K{=}8$ rollouts per query, so rollout generation cost is identical. The overhead comes from verifier inference and periodic DPO refresh (Table~\ref{tab:compute}). The verifier runs as a separate vLLM server, so its latency partially overlaps with policy gradient computation, and DPO refresh occurs only every $100$ steps. Overall, \modelname adds approximately $3$ hours ($3.6\%$) over standard GRPO.

\section{Prioritized Replay and Off-Policy Correction}
\label{app:replay}

\paragraph{Buffer structure.}
The prioritized replay buffer maintains a dictionary of the highest-priority entry seen per \texttt{prompt\_id}, backed by a max-heap for efficient capacity management.
Capacity is 4{,}096 entries.
Each entry stores the priority scalar, mode string, hardness estimate, and a timestamp.

\paragraph{Priority definition.}
After each rollout is scored, the buffer is updated with:
\begin{equation}
  p = |S_{\mathrm{verifier}}| + (1 - R_{\mathrm{ans}}),
\end{equation}
where $S_{\mathrm{verifier}}$ is the verifier score (before final clipping) and $R_{\mathrm{ans}}$ is the answer reward.
High-priority results from: (a) large verifier signal magnitude (indicating a grounding-important example), and (b) incorrect answers (indicating a still-challenging example).
Entries are only updated when the new priority exceeds the stored priority for that \texttt{prompt\_id}.

\paragraph{Replay bonus shaping.}
At reward-computation time, the stored priority for the current \texttt{prompt\_id} is looked up and added as a shaping bonus to the verifier reward:
\begin{equation}
  \Delta R_{\mathrm{replay}} = \gamma_r \cdot \min\!\left(\frac{p_{\mathrm{old}}}{2},\, 1.0\right), \quad \gamma_r = 0.15.
\end{equation}
This maintains gradient pressure on historically difficult prompts throughout training.

\paragraph{Off-policy importance sampling.}
When replay samples from earlier in training are mixed into a fresh batch, the policy has shifted since those completions were generated.
We correct for this with clipped importance-sampling weights:
\begin{equation}
  w_{\mathrm{IS}} = \frac{\pi_\theta(y|x)}{\pi_{\theta_{\mathrm{old}}}(y|x)}, \qquad
  \tilde{A}_t = \mathrm{clip}(w_{\mathrm{IS}},\, w_{\min},\, w_{\max}) \cdot A_t.
\end{equation}
Clipping prevents large IS weights from destabilizing training when the policy has moved substantially from the behavior that generated the replay sample.

\section{Failure Modes}
\label{app:limitations}

We identified the following failure modes through qualitative analysis of rollouts and mode distribution statistics across benchmarks:

\begin{itemize}
  \item \textbf{Mode collapse on uniformly easy batches.} When a mini-batch contains mostly easy queries (high $\hat{p}_{\mathrm{solved}}$), the adaptive length penalty (ALP) suppresses all reasoning modes similarly, reducing variance in the utility signal. DAPO's dynamic resampling partially mitigates this by discarding low-variance rollout groups.
  \item \textbf{Over-deliberation on difficult queries.} On genuinely difficult examples, \tagDeep occasionally produces excessively long reasoning traces with limited accuracy gains. The cost penalty reduces this behavior but does not fully eliminate it, since \tagDeep has no hard word cap.
  \item \textbf{Residual misrouting on Social and Multimodal Hint Reasoning queries.}
  On Video-Holmes~\citep{videoholmes}, the selected mode agrees with an independent GPT-5.5 mode judgment on $72\%$ of queries (Table~\ref{tab:holmes_routing}). The remaining disagreements concentrate in Social Reasoning and Multimodal Hint Reasoning queries, which appear to need only shallow single-hop inference but require evidence from multiple frames, so the policy can under-invoke deeper modes (see Figure~\ref{fig:qual_1}). A finer-grained mode space could close this gap.
\end{itemize}

\section{\texorpdfstring{Limitations}{Limitations}}
\label{sec:limitations}
\modelname judges reasoning quality with a learned verifier rather than human judgments, and the verifier may drift on out-of-distribution queries. The slot-based rollouts also spend computation on single-depth tasks such as pure classification. All experiments use Qwen-family backbones, and \tagDeep has the least SFT data because annotated deliberative traces are scarce (Appendix~\S\ref{app:data_construction}).

\section{\texorpdfstring{Broader Impacts}{Broader Impacts}}
\label{app:broader_impacts}
\modelname reduces inference cost by avoiding unnecessary reasoning, which lowers the energy use and deployment cost of capable multimodal models. Shorter responses also reduce latency, which makes multimodal assistants more practical in interactive settings and lowers the compute barrier for researchers and organizations with limited resources. The grounded traces cite objects, regions, and timestamps, so users can check the visual evidence behind an answer instead of taking it on trust. Avoiding unnecessary reasoning on perceptual queries also reduces hallucinated intermediate steps, which matters for applications such as assistive tools for blind and low-vision users and question answering over educational and scientific videos.

\section{Qualitative Analysis}
\label{app:qualitative}

\begin{figure*}[!t]
\centering
\includegraphics[width=\textwidth]{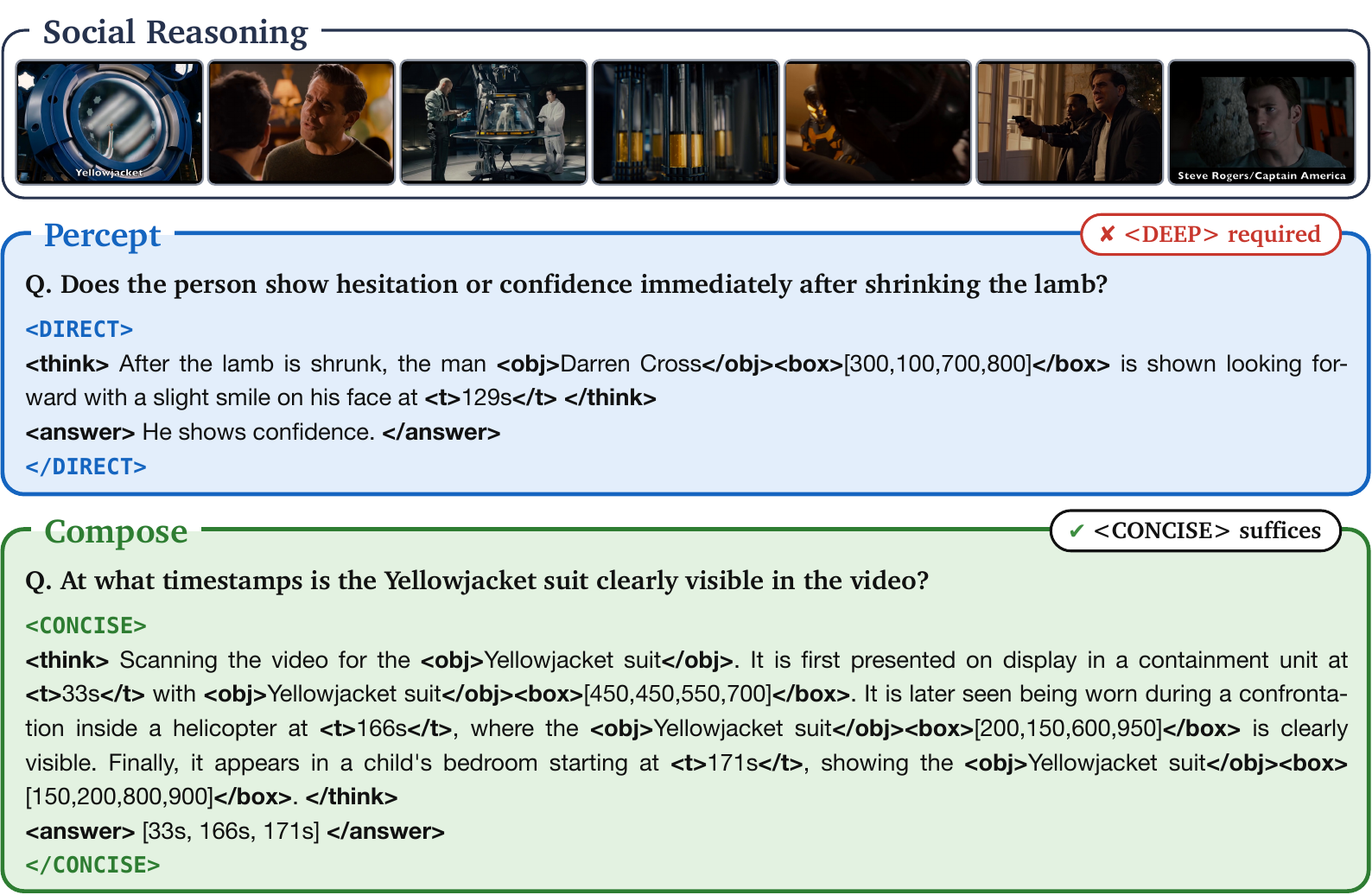}
\caption{\textbf{Social reasoning failure: \tagDeep under-invoked.}
Evaluating Darren Cross's emotional state after shrinking the lamb (top)
requires multi-frame deliberation across posture, gaze, and reaction cues, but the policy routes to \tagDirect and commits to a single-frame conclusion.
Grounding tags (\texttt{<obj>}, \texttt{<box>}, \texttt{<t>}) appear inside the \tagDirect block, signaling routing instability.
Locating the Yellowjacket suit across timestamps (bottom) is a sequential
scan well matched to \tagConcise, which correctly chains three grounded
appearances.}
\label{fig:qual_1}
\end{figure*}

\begin{figure*}[!t]
\centering
\includegraphics[width=\textwidth]{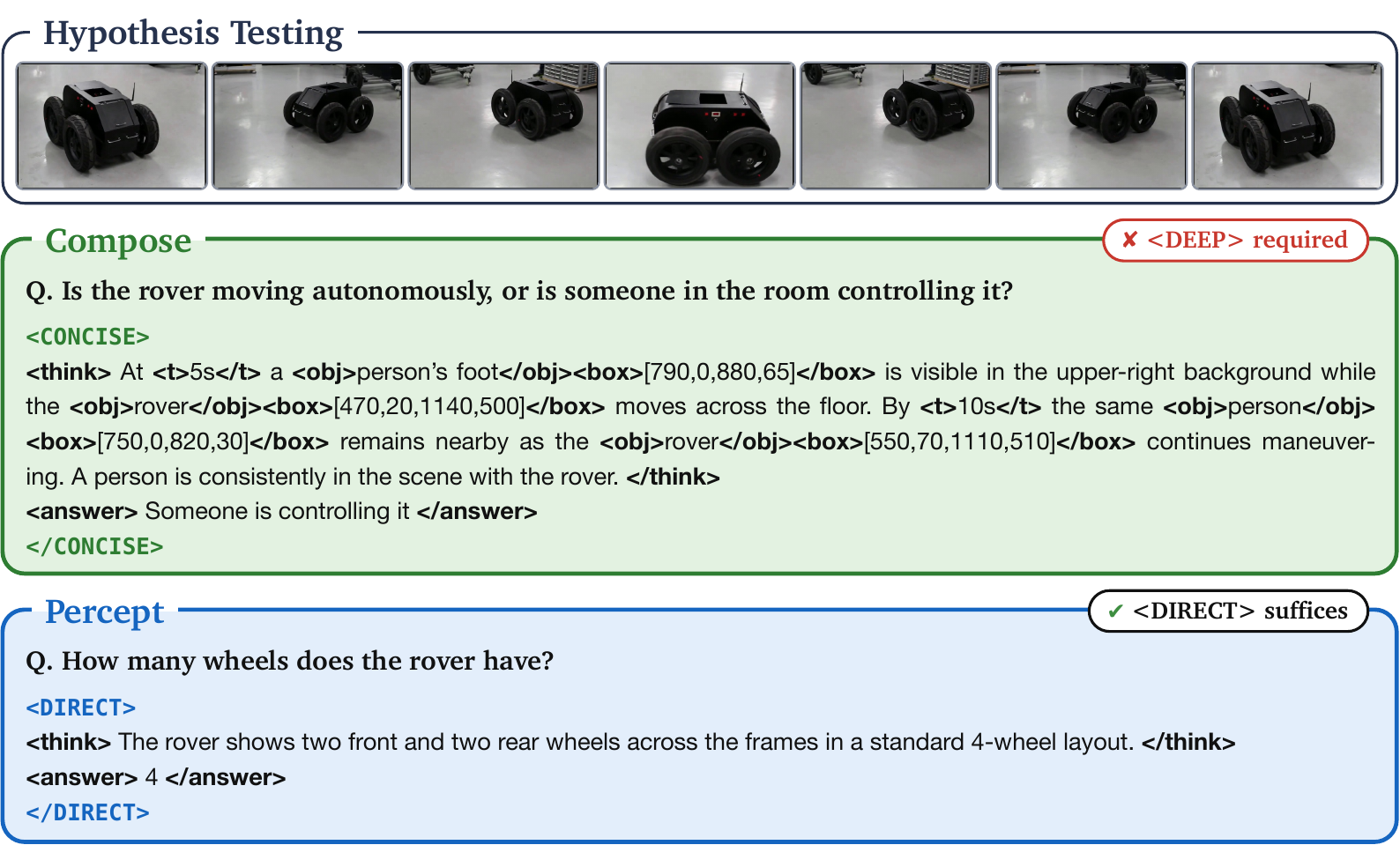}
\caption{\textbf{Compositional-surface query requiring hypothesis testing;
correct single-frame recognition.} Determining whether the rover moves autonomously (top)
requires falsifying the autonomous hypothesis across behavioral cues (controller
presence, motion smoothness, gaze tracking). The policy routes to \tagConcise
and chains co-location observations without testing any of those alternatives.
\tagDeep was needed. Counting wheels (bottom) is single-frame recognition with no
competing hypotheses. \tagDirect is cost-optimal.}
\label{fig:qual_2}
\end{figure*}

\begin{figure*}[!t]
\centering
\includegraphics[width=\textwidth]{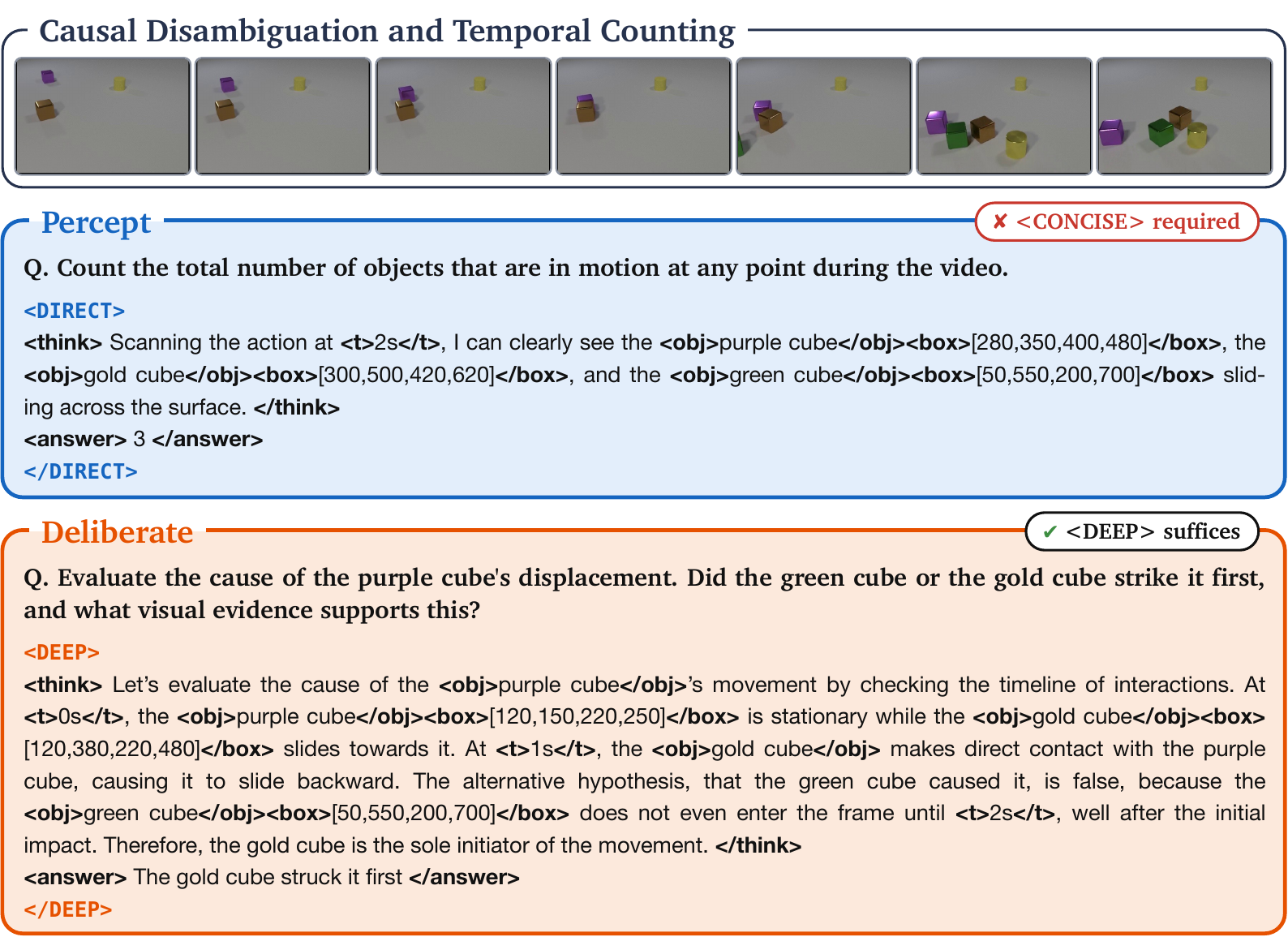}
\caption{\textbf{Failure on temporal counting; correct causal hypothesis testing.}
The counting query (top) demands a running
tally across the full video. The policy routes to \tagDirect, inspects
a single frame, and reports $3$ objects instead of the correct $4$, missing
the gold cylinder, which moves only in other segments.
The causality query (bottom) requires falsifying the green-cube hypothesis.
\tagDeep correctly establishes the green cube does not enter the frame
until after the initial impact, then confirms gold-cube contact via
timestamped bounding boxes.}
\label{fig:qual_3}
\end{figure*}

\begin{figure*}[!t]
\centering
\includegraphics[width=\textwidth]{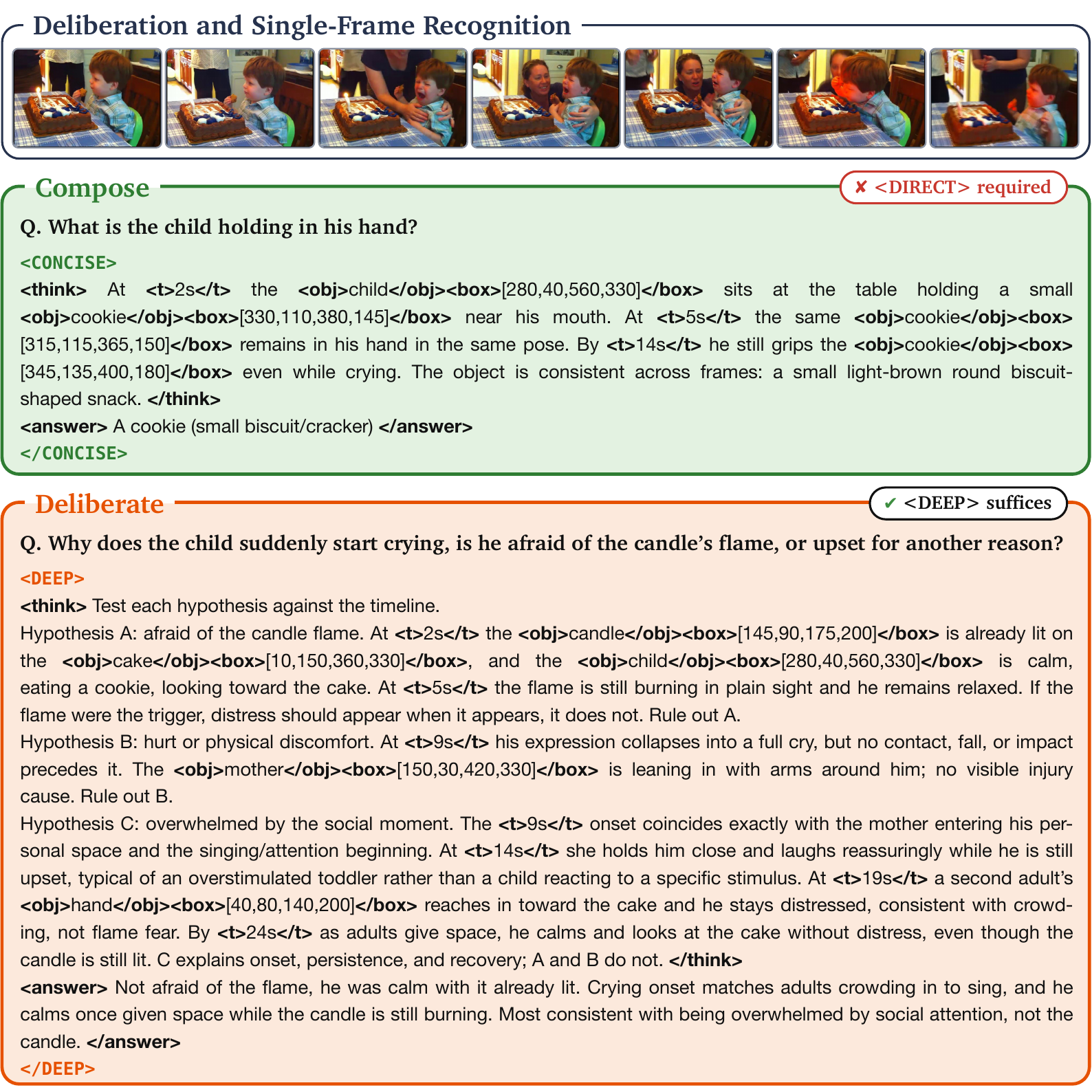}
\caption{\textbf{Over-routing single-frame recognition; correct deliberative
causal reasoning.} The object-identification query (top) is answerable
from any single frame. \tagConcise chains three cross-frame
confirmations when \tagDirect's one-step extraction was sufficient,
wasting tokens with no accuracy gain. The distress query (bottom) presents three competing
explanations (flame fear, physical discomfort, social overwhelm).
\tagDeep tests each with timestamped grounding and rules out the
first two before confirming social overwhelm via co-occurring onset and
recovery cues.}
\label{fig:qual_4}
\end{figure*}

Figures~\ref{fig:qual_1}--\ref{fig:qual_4} show representative correct and incorrect mode selections. Each figure pairs one success case with one failure case on the same video.

\paragraph{Social reasoning failure (Figure~\ref{fig:qual_1}).}
Assessing Darren Cross's emotional state immediately after shrinking the lamb requires evaluating posture, gaze, and reaction across multiple frames to rule out a superficial interpretation of confidence. This is the hypothesis-testing query for \tagDeep. The policy routes to \tagDirect, commits to a single-frame observation, and
produces an incorrect conclusion. The response also generates grounding tags (\texttt{<obj>}, \texttt{<box>}, \texttt{<t>}) inside \tagDirect output, a signal of routing instability.
This failure occurs in the Social Reasoning subtask of Video-Holmes (\S\ref{app:limitations}), where social and emotional queries may appear to require only single-frame recognition, leading the cost-aware policy to under-select \tagDeep. The Yellowjacket timestamp query (bottom) is a sequential scan. \tagConcise
correctly chains three grounded appearances across the video.

\paragraph{Compositional-surface query requiring hypothesis testing (Figure~\ref{fig:qual_2}).}
Determining whether the rover is autonomous or operator-controlled requires
falsifying the autonomous hypothesis across multiple behavioral dimensions:
controller presence, motion smoothness, and gaze tracking.
The policy routes to \tagConcise, which linearly chains co-location observations without evaluating competing interpretations. \tagDeep is required because the question cannot be resolved through evidence accumulation alone. It requires actively rejecting alternative explanations. In contrast, the wheel-count query (bottom) involves no competing hypotheses and can be answered from any clear frame, making \tagDirect the most cost-efficient choice.

\paragraph{Temporal counting and causal disambiguation (Figure~\ref{fig:qual_3}).}
The counting query demands a cumulative scan: the correct answer ($4$ objects:
gold cube, purple cube, green cube, and gold cylinder) requires a running tally
over the full video, not a snapshot. The policy routes to \tagDirect, inspects
one frame, and reports $3$, missing the gold cylinder, which is in motion only in other
segments. Counting tasks have the surface form of perceptual questions, but their
temporal structure requires \tagConcise.
The causality query (bottom) explicitly requires falsifying a competing hypothesis, the
textbook query for \tagDeep. The policy correctly establishes that the green
cube does not enter the frame until after the initial impact, eliminating that
hypothesis, and confirms gold-cube contact with timestamped bounding boxes.

\paragraph{Over-routing single-frame recognition, correct deliberation (Figure~\ref{fig:qual_4}).}
The object-identification query is answerable from any single frame.
\tagConcise chains three cross-frame confirmations of the same cookie.
By Eq.~\ref{eq:utility}, $\mathrm{Ans}(\tagDirect) \approx
\mathrm{Ans}(\tagConcise)$ while $\mathrm{Cost}(\tagConcise) >
\mathrm{Cost}(\tagDirect)$, so \tagDirect should win on utility.
This is a milder instance of the perception-task over-reasoning noted in
\S\ref{sec:experiments}: cost pressure is not yet sharp enough to suppress
\tagConcise on queries that require no temporal aggregation.
The distress query (bottom) presents three mutually exclusive explanations that must each be
tested. \tagDeep tests flame fear (child is calm while the flame
is already lit), physical discomfort (no contact precedes the cry), and social
overwhelm (onset co-occurs with crowding; calming follows once space is given).
Each step is tied to a timestamped observation, not merely listed.

\clearpage

\clearpage

\clearpage
\section{Prompt Templates}
\label{app:prompts}
Figures~\ref{fig:prompt_direct}--\ref{fig:prompt_verifier} show the complete prompt text for each mode and for the verifier.
\begin{figure}[!h]
\centering
\begin{tcolorbox}[
  colback=gray!12,
  colframe=black,
  coltitle=white,
  fonttitle=\bfseries\small,
  title={DIRECT Mode Training Prompt},
  left=3mm, right=3mm, top=1mm, bottom=1mm,
  boxrule=0.6pt
]
\begin{lstlisting}[basicstyle=\footnotesize\ttfamily,columns=flexible,breaklines=true,breakatwhitespace=true,frame=none,aboveskip=0pt,belowskip=0pt]
You are answering a multimodal question in DIRECT mode.

Assigned mode: DIRECT

Output format (must follow exactly):
<DIRECT>
<think>
Brief internal reasoning with minimal verification only (<=60 words).
</think>
<answer>
Final answer.
</answer>
</DIRECT>

Rules:
- Use only DIRECT wrapper tags.
- Exactly one <think> and one <answer>.
- No text outside <DIRECT>...</DIRECT>.
- Be decisive and concise.
- Use only evidence visible in the provided input (video/image/text).
- If options are provided (e.g., multiple choice), output the answer
  in the expected option style.
\end{lstlisting}
\end{tcolorbox}
\caption{\textbf{DIRECT mode prompt.} Targets fast, minimal-verification responses ($\leq$60 \texttt{<think>} words). Used for controlled slots 1--2 during training.}
\label{fig:prompt_direct}
\end{figure}

\begin{figure}[!h]
\centering
\begin{tcolorbox}[
  colback=gray!12,
  colframe=black,
  coltitle=white,
  fonttitle=\bfseries\small,
  title={CONCISE Mode Training Prompt},
  left=3mm, right=3mm, top=1mm, bottom=1mm,
  boxrule=0.6pt
]
\begin{lstlisting}[basicstyle=\footnotesize\ttfamily,columns=flexible,breaklines=true,breakatwhitespace=true,frame=none,aboveskip=0pt,belowskip=0pt]
You are answering a multimodal question in CONCISE mode.

Assigned mode: CONCISE

Output format (must follow exactly):
<CONCISE>
<think>
Compact reasoning with key verification/falsification checks (<=200 words).
</think>
<answer>
Final answer.
</answer>
</CONCISE>

Rules:
- Use only CONCISE wrapper tags.
- Exactly one <think> and one <answer>.
- No text outside <CONCISE>...</CONCISE>.
- Ground claims in observed evidence; include only key checks.
- Keep reasoning compact and useful.
- Include grounding tags inside <think> when useful:
  <obj>object_name</obj><box>[x1,y1,x2,y2]</box><t>time_in_seconds</t>
- Do not output tags in <answer>.
- If options are provided, output in the expected answer style.
\end{lstlisting}
\end{tcolorbox}
\caption{\textbf{CONCISE mode prompt.} Targets compact grounded reasoning ($\leq$200 \texttt{<think>} words) with optional spatio-temporal grounding tags. Used for controlled slots 3--4.}
\label{fig:prompt_concise}
\end{figure}

\begin{figure}[t]
\centering
\begin{tcolorbox}[
  colback=gray!12,
  colframe=black,
  coltitle=white,
  fonttitle=\bfseries\small,
  title={DEEP Mode Training Prompt},
  left=3mm, right=3mm, top=1mm, bottom=1mm,
  boxrule=0.6pt
]
\begin{lstlisting}[basicstyle=\footnotesize\ttfamily,columns=flexible,breaklines=true,breakatwhitespace=true,frame=none,aboveskip=0pt,belowskip=0pt]
You are answering a multimodal question in DEEP mode.

Assigned mode: DEEP

Output format (must follow exactly):
<DEEP>
<think>
Detailed verification/falsification reasoning with assumptions, alternatives,
counterexamples, and failure cases; grounded in observed evidence (<=1000 words).
</think>
<answer>
Final answer.
</answer>
</DEEP>

Rules:
- Use only DEEP wrapper tags.
- Exactly one <think> and one <answer>.
- No text outside <DEEP>...</DEEP>.
- Reason step-by-step and verify competing hypotheses before concluding.
- Avoid unsupported claims; do not invent details.
- Include explicit evidence references (objects/actions/time cues) when relevant.
- Include grounding tags inside <think> when useful:
  <obj>object_name</obj><box>[x1,y1,x2,y2]</box><t>time_in_seconds</t>
- Do not output tags in <answer>.
- If options are provided, output in the expected answer style.
\end{lstlisting}
\end{tcolorbox}
\caption{\textbf{DEEP mode prompt.} Targets thorough hypothesis-falsification reasoning ($\leq$1000 \texttt{<think>} words) with grounding evidence. Used for controlled slots 5--6.}
\label{fig:prompt_deep}
\end{figure}

\begin{figure}[t]
\centering
\begin{tcolorbox}[
  colback=gray!12,
  colframe=black,
  coltitle=white,
  fonttitle=\bfseries\small,
  title={ADAPTIVE Inference Prompt},
  left=3mm, right=3mm, top=1mm, bottom=1mm,
  boxrule=0.6pt
]
\begin{lstlisting}[basicstyle=\footnotesize\ttfamily,columns=flexible,breaklines=true,breakatwhitespace=true,frame=none,aboveskip=0pt,belowskip=0pt]
You are in ADAPTIVE inference mode.
Your job is to choose the cheapest reliable reasoning style for this
question: DIRECT, CONCISE, or DEEP.

Mode definitions:
- DIRECT: Fastest, minimal verification, shortest reasoning.
- CONCISE: Moderate cost, compact but grounded checks.
- DEEP: Highest cost, thorough verification with alternatives/failure-case analysis.

Decision policy:
1) Estimate task difficulty and ambiguity from the input.
2) Prefer DIRECT when confidence is high and evidence is clear.
3) Use CONCISE when moderate verification is needed.
4) Use DEEP only when uncertainty, multi-step reasoning, conflicting evidence,
   or strong grounding demands require it.

Then answer using exactly one corresponding wrapper format:
<DIRECT><think>...</think><answer>...</answer></DIRECT>
<CONCISE><think>...</think><answer>...</answer></CONCISE>
<DEEP><think>...</think><answer>...</answer></DEEP>

Rules:
- Do NOT use <ADAPTIVE> wrapper.
- Exactly one <think> and one <answer>.
- No text outside the chosen wrapper.
- Ground claims in provided evidence only.
- If options are provided, output in the expected answer style.
\end{lstlisting}
\end{tcolorbox}
\caption{\textbf{ADAPTIVE mode inference prompt.} Instructs the model to select the cheapest reliable mode via a cost-first decision policy. Used for adaptive slots 7--8 at both training and inference time.}
\label{fig:prompt_adaptive}
\end{figure}

\begin{figure}[t]
\centering
\begin{tcolorbox}[
  colback=gray!12,
  colframe=black,
  coltitle=white,
  fonttitle=\bfseries\small,
  title={Verifier System Prompt},
  left=3mm, right=3mm, top=1mm, bottom=1mm,
  boxrule=0.6pt
]
\begin{lstlisting}[basicstyle=\footnotesize\ttfamily,columns=flexible,breaklines=true,breakatwhitespace=true,frame=none,aboveskip=0pt,belowskip=0pt]
You are a strict multimodal verifier for adaptive reasoning outputs.
Evaluate whether the candidate response is correct, grounded, and
human-aligned, and whether the chosen reasoning mode is appropriate.

Inputs: question, candidate completion (with mode wrapper), ground-truth
answer, and media context (video/image) when available.

Return ONLY valid JSON with this schema:
{
  "temporal_grounding": float,        // 0.0 to 1.0
  "spatial_grounding":  float,        // 0.0 to 1.0
  "human_alignment":    float,        // 0.0 to 1.0
  "preferred_mode":     "DIRECT|CONCISE|DEEP",
  "reasoning_text_ok":  0 or 1,
  "span_scores":        [float, ...]  // optional; each in [-1.0, 1.0]
}

Scoring rubric:
1) temporal_grounding: 1.0 = clear time-localized evidence; 0.5 = partial;
   0.0 = no temporal grounding where required, or wrong temporal claims.
2) spatial_grounding: 1.0 = concrete object/region anchoring; 0.5 = partial;
   0.0 = no spatial support or contradictory visual assertions.
3) human_alignment: evaluate helpfulness, coherence, non-contradiction, and
   instruction-following; penalize fabricated evidence, self-contradiction,
   unsafe or unreadable reasoning.
4) preferred_mode: cheapest mode judged reliable for this sample.
   Use DIRECT < CONCISE < DEEP as cost prior.
5) reasoning_text_ok: 1 if <think> contains meaningful natural-language
   reasoning; 0 if empty, repetitive, or non-informative.
6) span_scores: per-span scores in [-1,1]; positive = useful
   evidence-bearing span; negative = filler, hedging, or hallucination.

General rules:
- Be strict and calibration-aware.
- Use media evidence when available; for text-only tasks, use textual logic.
- Do not output any explanation outside JSON.
- Output must be machine-parseable JSON only.
\end{lstlisting}
\end{tcolorbox}
\caption{\textbf{Verifier system prompt.} The verifier (a Qwen3-VL model) receives each candidate completion and returns a structured JSON signal covering temporal grounding, spatial grounding, human alignment, preferred mode, reasoning quality, and span-level scores.}
\label{fig:prompt_verifier}
\end{figure}

\clearpage

\end{document}